\documentclass[10pt,twocolumn,letterpaper]{article}

\usepackage{iccv}      % To produce the REVIEW version
\usepackage[utf8]{inputenc} % allow utf-8 input
\usepackage[T1]{fontenc}    % use 8-bit T1 fonts
\usepackage{url}            % simple URL typesetting
\usepackage{booktabs}       % professional-quality tables
\usepackage{amsfonts} 
\usepackage{comment}% blackboard math symbols
\usepackage{nicefrac}       % compact symbols for 1/2, etc.
\usepackage{microtype}      % microtypography
\usepackage{xcolor}         % colors
\usepackage{amsmath,bm}
\usepackage{float}
\usepackage{multirow, graphicx}
\usepackage{graphicx}
\usepackage{tabularx, makecell, multirow} 
\usepackage{wrapfig}
\usepackage{amsmath, amssymb, amsthm, geometry, mathtools}
\usepackage{graphicx}
\usepackage{perl_acronyms}

\usepackage{amsfonts}
\usepackage{subcaption}    % for subfigures

\usepackage{amsmath,amssymb}

\usepackage[dvipsnames]{xcolor}
\usepackage{xcolor,colortbl}
\usepackage{multirow, makecell}
\definecolor{Gray}{gray}{0.50}
\newcolumntype{g}{>{\columncolor{Gray}}c}
\definecolor{ffe1da}{RGB}{255,225,218}
\definecolor{F7E0D5}{RGB}{247,224,213}
\definecolor{darkF7E0D5}{RGB}{209,154,128}
\colorlet{Light}{White!0!F7E0D5}

\colorlet{tabfirst}{Green!25}
\definecolor{tabthird}{rgb}{1, 0.85, 0.7}
\definecolor{tabsecond}{rgb}{1, 0.96, 0.7}

\definecolor{tabfirst}{rgb}{0.79, 0.92, 1.0} % Light Blue with High Saturation
\definecolor{tabsecond}{rgb}{1.0, 0.86, 0.86} % Light Red with High Saturation
\definecolor{tabthird}{rgb}{0.9, 0.8, 1.0} % Light Purple with High Saturation

\definecolor{cvprblue}{rgb}{0.21,0.49,0.74}
\definecolor{mypurple}{rgb}{0.4549,0.145,0.99}
\usepackage[pagebackref,breaklinks,colorlinks,allcolors=cvprblue]{hyperref}

\def\paperID{1468} % *** Enter the Paper ID here
\def\confName{ICCV}
\def\confYear{2025}

\title{Towards Ambiguity-Free Spatial Foundation Model:\\ Rethinking and Decoupling Depth Ambiguity}

\usepackage{xcolor}
\usepackage{calc}
\definecolor{tabfirst}{rgb}{0.95, 0.99, 1.0}   % Lightest Blue
\definecolor{tabsecond}{rgb}{0.82, 0.95, 1.0}  % Medium Blue
\definecolor{tabthird}{rgb}{0.75, 0.90, 1.0}   % Darkest Blue

\definecolor{tabfirst}{rgb}{0.92, 0.97, 1.0}   % Very Light Blue
\definecolor{tabsecond}{rgb}{0.85, 0.90, 0.98}  % Light Blue
\definecolor{tabthird}{rgb}{0.75, 0.85, 0.99}   % Medium Blue

\colorlet{tabthird}{Green!30}  % Light Green (25% Green)
\colorlet{tabsecond}{Green!18}   %{rgb}{1, 0.96, 0.7}   % Very Light Yellow-Orange
\colorlet{tabfirst}{Green!5}  %{rgb}{1, 0.85, 0.7}    % Light Yellow-Orange

\colorlet{tabthird}{Green!25}  % Light Green (25% Green)
\definecolor{tabsecond}{rgb}{1, 0.96, 0.7}   % Very Light Yellow-Orange
\definecolor{tabfirst}{rgb}{1, 0.88, 0.72}    % Light Yellow-Orange
\colorlet{tabgreen}{Green!10}
\definecolor{tabfirst}{rgb}{0.92, 0.97, 1.0}   % Very Light Blue
\definecolor{tabsecond}{rgb}{0.85, 0.90, 0.98}  % Light Blue
\definecolor{tabthird}{rgb}{0.75, 0.85, 0.99}   % Medium Blue

\definecolor{tabfirst}{rgb}{0.79, 0.92, 1.0} % Light Blue with High Saturation
\definecolor{tabsecond}{rgb}{1.0, 0.86, 0.86} % Light Red with High Saturation
\definecolor{tabthird}{rgb}{0.65, 0.92, 0.87}

\definecolor{tabfirst}{rgb}{0.82, 0.89, 0.99} % Light Blue
\definecolor{tabsecond}{rgb}{0.98, 0.82, 0.81} % Light Red
\definecolor{tabthird}{rgb}{0.81, 0.92, 0.84} % Light Green

\author{
    $\text{Xiaohao Xu}^{1}$ \quad
    $\text{Feng Xue}^{1}$\quad
    $\text{Xiang Li}^{2}$\quad
    $\text{Haowei Li}^{1}$\quad 
    $\text{Shusheng Yang}^{3}$\\
    $\text{Tianyi Zhang}^{2}$\quad
    $\text{Matthew Johnson-Roberson}^{2}$\quad
    $\text{Xiaonan Huang}^{1}$\\ 
    $^{1}$University of Michigan, Ann Arbor \quad $^{2}$Carnegie Mellon University\quad $^{3}$New York University\ \\
    }
\begin{document}
\maketitle
\begin{abstract}

Depth ambiguity is a fundamental challenge in spatial scene understanding, especially in transparent scenes where single-depth estimates fail to capture full 3D structure. Existing models, limited to deterministic predictions, overlook real-world multi-layer depth. To address this, we introduce a paradigm shift from single-prediction to multi-hypothesis spatial foundation models.
We first present \texttt{MD-3k}, a benchmark exposing depth biases in expert and foundational models through multi-layer spatial relationship labels and new metrics.
To resolve depth ambiguity, we propose Laplacian Visual Prompting (LVP), a training-free spectral prompting technique that extracts hidden depth from pre-trained models via Laplacian-transformed RGB inputs. By integrating LVP-inferred depth with standard RGB-based estimates, our approach elicits multi-layer depth without model retraining.
Extensive experiments validate the effectiveness of LVP in zero-shot multi-layer depth estimation, unlocking more robust and comprehensive geometry-conditioned visual generation, 3D-grounded spatial reasoning, and temporally consistent video-level depth inference. Our benchmark and code will be available at \url{https://github.com/Xiaohao-Xu/Ambiguity-in-Space}. 
\end{abstract}

\section{{Introduction}}
\label{sec:introduction}

\vspace{-0.5mm}
Spatial understanding, the ability to derive structured 3D representations from sensory data, is fundamental to visual intelligence and autonomous systems.  Despite progress in both physical sensors and monocular depth estimation models~\cite{midas, midasv31, zoedepth, metric3d, zerodepth, patchfusion, marigold, depth_anything, depthfm, geowizard, unidepth, metric3dv2, yang2024depth} (see Fig.~\ref{fig:teaser}a\&b), a key challenge persists: \textbf{biased 3D spatial understanding under depth ambiguity}.

In real-world 3D scenes, factors such as transparency (see Fig.\ref{fig:teaser}c) break the assumption that each pixel corresponds to a unique depth value. For example, objects viewed through transparent surfaces like glass exhibit a range of plausible depths rather than a single fixed value. While state-of-the-art depth foundation models~\cite{yang2024depth,depth_anything} generalize well in unambiguous scenarios, they typically output only a single depth estimate, thereby ignoring inherent depth ambiguity. This limitation results in \textbf{biased, incomplete 3D representations} that undermine both generalization and reliability, especially in safety-critical applications requiring robust spatial reasoning.

To this end, we advocate a paradigm shift from single-prediction to {Multi-Hypothesis Spatial Foundation Models} (MH-SFMs). We posit that \textbf{true spatial intelligence demands explicitly modeling and resolving ambiguity} rather than forcing a biased single-depth output. To address this, we propose a unified framework that enables multi-layer depth estimation from a monocular image via a single, domain-agnostic foundation model (see Fig.~\ref{fig:teaser}d).

\begin{figure}[t!]
\centering
\setlength{\abovecaptionskip}{0.2cm}
\includegraphics[width=0.48\textwidth]{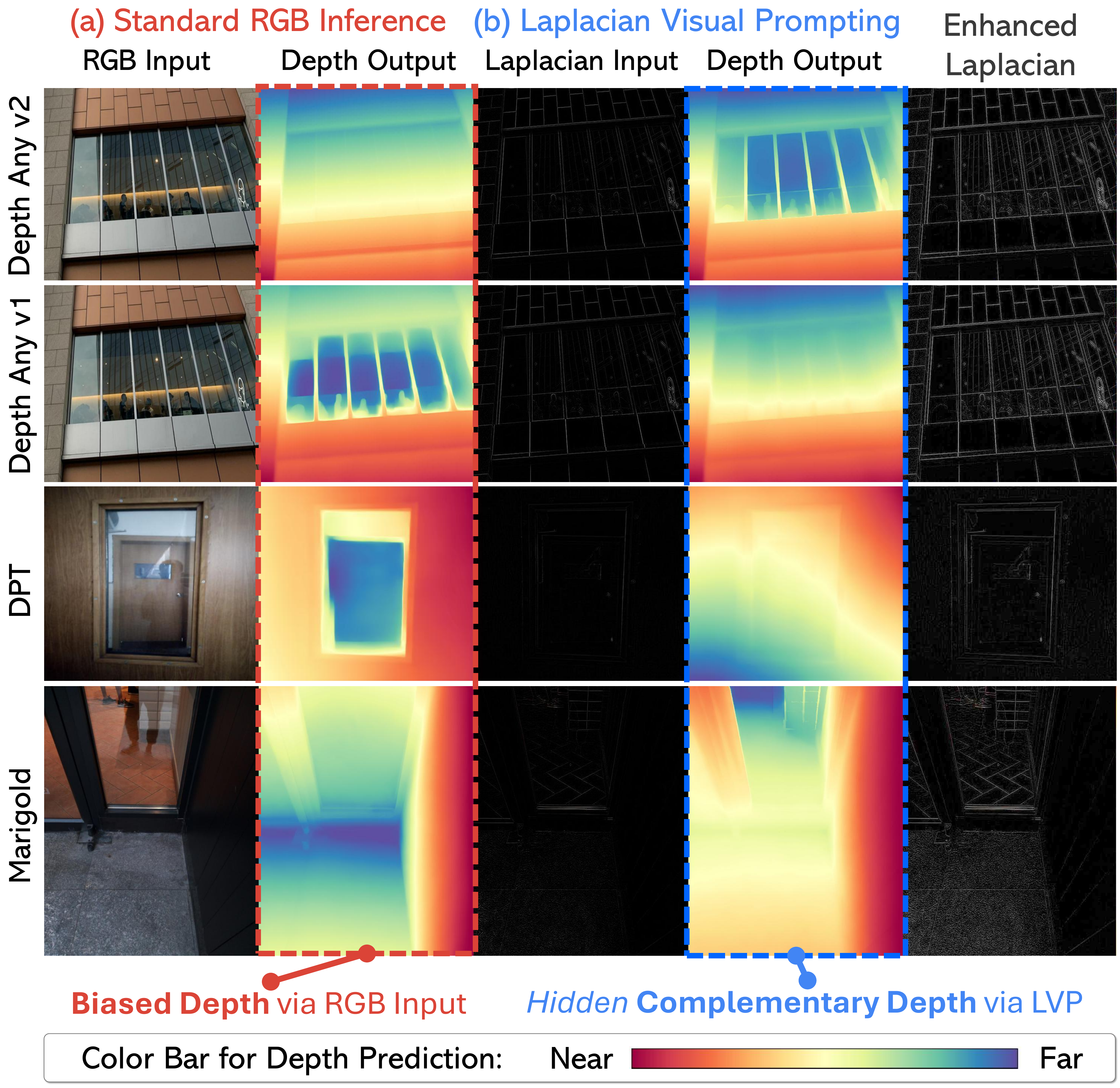}
\caption{\textbf{Unlocking \textit{hidden depth} with Laplacian Visual Prompting across diverse baselines~\cite{depth_anything,yang2024depth,marigold,dpt}.} Each case includes the RGB input, estimated depth from RGB, Laplacian input, estimated \textit{hidden depth} from Laplacian, and an enhanced Laplacian.  {Notice that depth maps from  RGB and LVP } \textbf{both capture \textit{plausible hypotheses}}: one for the \textbf{\textit{transparent}} surface (glass) and another for the \textbf{\textit{opaque}} object behind it.}
\label{fig:hidden_depth_1}\vspace{-3mm}
\end{figure}

To enable rigorous study of multi-layer spatial relationships under depth ambiguity, we introduce \texttt{MD-3k}, a benchmark featuring explicit labels for multilayer spatial relationships that goes beyond traditional single-depth metrics. Our analysis reveals that existing models exhibit significant depth biases under standard RGB input—some favoring nearer surfaces, others preferring farther ones (see Fig.~\ref{fig:hidden_depth_1}a)—highlighting the limitations of the conventional single-depth prediction paradigm.

Next, we introduce {Laplacian Visual Prompting} (LVP), a training-free spectral prompting technique for 3D spatial decoupling.  LVP draws inspiration from prompting techniques in NLP~\cite{wei2022chain} and visual prompting~\cite{bar2022visual,hojel2025finding,bai2024sequential}.  At its core, LVP applies the discrete Laplacian operator, a fundamental second-order difference operator, to the RGB image input.  This operation generates high-frequency visual prompts that highlight regions of rapid intensity change, effectively exposing latent spatial knowledge within pre-trained depth models.  Integrating depth maps from LVP and RGB inputs enables multi-hypothesis depth estimation without retraining, revealing pre-trained models' latent ability to disentangle multi-layered 3D structures.  We demonstrate LVP's effectiveness on the \texttt{MD-3k} benchmark, showing that it uncovers hidden depth (see Fig.~\ref{fig:hidden_depth_1}b) and  mitigates inherent depth biases.  Further analysis using LVP explores the scaling laws of spatial understanding under ambiguous and non-ambiguous scenes, and identifies key challenges in resolving multi-layer spatial relationships.

%We demonstrate LVP's effectiveness on the \texttt{MD-3k} benchmark, showing that it uncovers hidden depth (see Fig.~\ref{fig_hidden_depth_1}b) and mitigates inherent depth biases. \textbf{Furthermore, our scaling law analysis reveals that simply increasing model size does not guarantee improved spatial understanding, particularly in ambiguous scenes, underscoring the need for more nuanced approaches beyond naive scaling.} Further analysis using LVP explores the scaling laws of spatial understanding under ambiguous and non-ambiguous scenes, and identifies key challenges in resolving multi-layer spatial relationships.

Finally, we demonstrate the practical benefits of LVP’s multi-hypothesis depth estimation, enabling flexible geometry-conditioned visual generation~\cite{controlnet}, including realistic 3D re-synthesis of transparent structures for ambiguous scenes, consistent multi-layer depth estimation in real-world videos, and robust 3D spatial reasoning in multi-modal LLMs. These results highlight the potential of LVP in advancing spatial intelligence.
%Finally, we demonstrate the practical benefits of multi-hypothesis depth estimation with LVP. This approach enables flexible geometry-conditioned visual generation~\cite{controlnet} and realistic re-synthesis of 3D scenes with transparent structures, while also consistently revealing hidden depth in real-world videos and enhancing 3D spatial reasoning in multi-modal large language models. Together, these results highlight the potential of MH-SFMs and LVP to advance reliable spatial intelligence.

\textbf{In summary, our main contributions are:}
\begin{itemize}
    \item We rethink spatial ambiguity in real-world 3D scenes and reformulate domain-agnostic, (\textit{i.e.}, foundational) depth estimation as {multi-hypothesis inference}.
    \item We introduce \texttt{\texttt{\texttt{MD-3k}}}, a new benchmark to evaluate multilayer spatial understanding and model biases.
    \item We analyze existing models across diverse architecture, training schema, and model size on \texttt{\texttt{\texttt{MD-3k}}} and reveal {different depth biases} under ambiguity.
    \item We propose {{Laplacian Visual Prompting} (LVP)}, a {training-free prompting method}, to facilitate {multi-hypothesis depth} estimation from pre-trained models.
    \item We validate LVP's effectiveness in revealing multi-layer depth and depth bias control. We also demonstrate its benefits for geometry-conditioned visual generation, spatial reasoning, and video-level inference.
\end{itemize}

\begin{figure}[t!]
    \centering
    \setlength{\abovecaptionskip}{0.2cm}
    \includegraphics[width=0.48\textwidth]{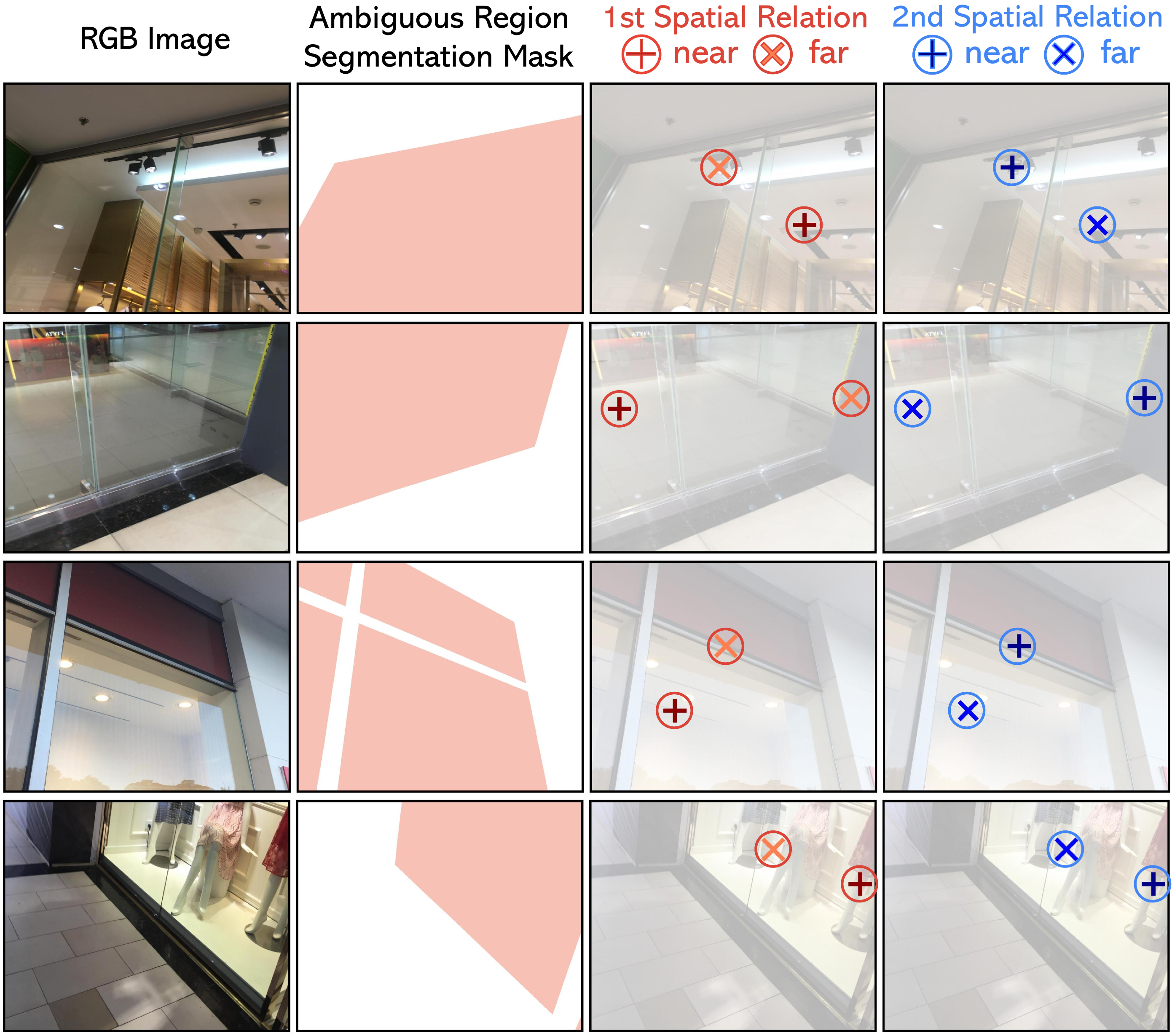}
\caption{\textbf{\texttt{MD-3k} benchmark for evaluating multi-layer spatial relationships.} Example images feature annotated ambiguous region masks and sparse point pairs with multi-layer spatial labels. The first and second spatial relation columns show ground truth near/far annotations (red and blue markers, respectively). The top three rows depict \textit{\textbf{reverse}} relationships, while the bottom row shows a \textit{\textbf{same}} relationship between layers.}
    \label{fig:benchmark}\vspace{-3mm}
\end{figure}
\section{Related Work}

\noindent\textbf{Monocular depth estimation (MDE)}. MDE has evolved from early domain-specific depth estimation \cite{eigen2014depth,fu2018deep,adabins}, constrained by dataset-specific training (\textit{e.g.}, NYU \cite{nyud}, KITTI \cite{kitti}), to more generalizable domain-agnostic approaches, pushing forward the frontier towards generic depth foundation models. Recent methods exploit Stable Diffusion \cite{sd} for fine-grained depth prediction \cite{marigold, depthfm, geowizard}. MiDaS \cite{midas, dpt, midasv31} and Metric3D \cite{metric3d} rely on labeled data, while Depth Anything V1 \cite{depth_anything} and V2~\cite{yang2024depth} enhance robustness through large-scale and pseudo-labeled training.
Despite these advances, \textit{\textbf{existing monocular depth foundation models estimate only single-layer depth, struggling with multi-layer spatial ambiguities.}} To address this, we redefine depth estimation in a domain-agnostic setting as a multi-hypothesis problem, using Laplacian Visual Prompting to disentangle depth layers in ambiguous visual contexts. %MDE has progressed from early methods focused on in-domain metric accuracy 

\vspace{0.5mm}
\noindent\textbf{Visual prompting (VP)}. Inspired by prompt-based adaptation in NLP \cite{brown2020language}, VP \cite{bahng2022visual,bai2024sequential} enables pre-trained vision models to be adapted via input-space manipulation.  VP has been successfully applied to vision-language models \cite{bahng2022visual,singha2023ad,wasim2023vita}, with further improvements through joint text-visual optimization \cite{khattak2022maple,wang2024vilt}.
In addition, VP has been explored for black-box model adaptation \cite{tsai2020transfer}, cross-domain transfer \cite{chen2021adversarial, neekhara2022cross}, and adversarial robustness \cite{chen2022visual}. While VP research has primarily focused on semantic understanding tasks, \textbf{\textit{its potential for 3D spatial decoupling and comprehension remains largely unexplored.}} To address this gap, we introduce Laplacian Visual Prompting, which facilitates training-free spatial 3D  decoupling through multi-hypothesis depth estimation.

\begin{figure}[t!]
    \centering
    \setlength{\abovecaptionskip}{0.2cm}
    \includegraphics[width=0.48\textwidth]{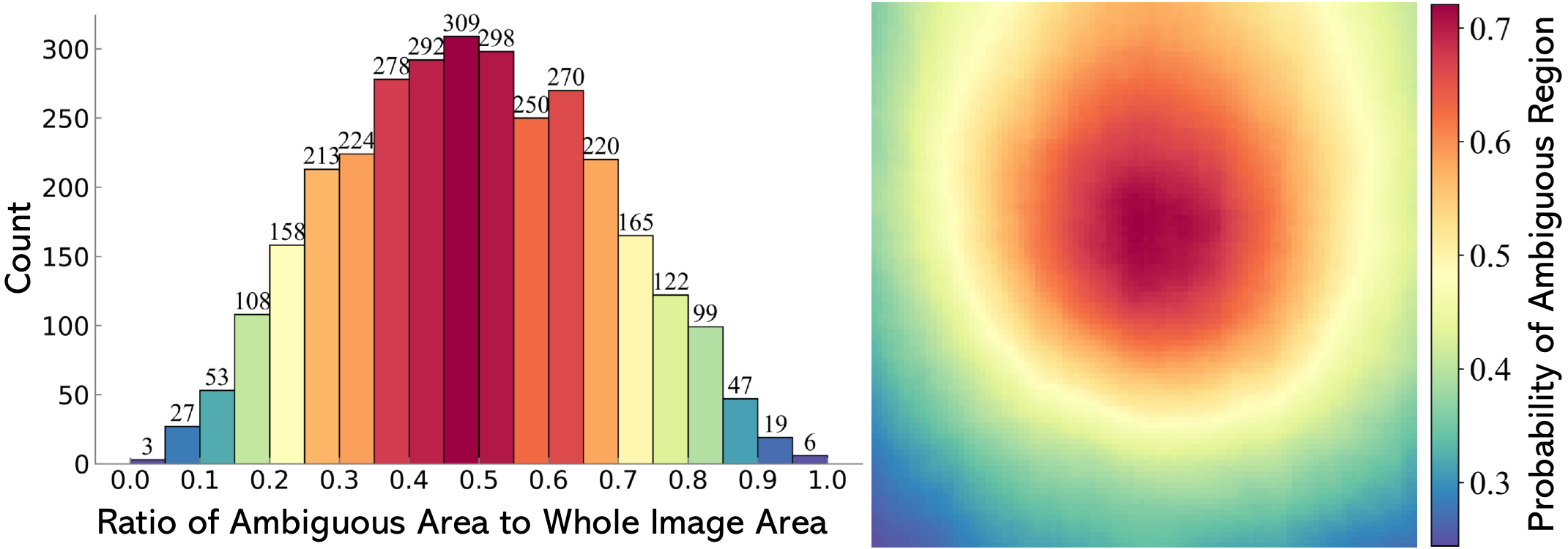}
\caption{\textbf{Statistics of ambiguous regions in the \texttt{\textbf{\texttt{MD-3k}}} benchmark.} \textcolor{black}{Ratio of ambiguous regions to the whole image (\textbf{Left}) and spatial distribution of ambiguity regions (\textbf{Right}) }}
    \label{fig:benchmark-distribution} \vspace{-1mm}
\end{figure}
\section{Multi-Hypothesis Depth Estimation}

Monocular depth estimation in complex 3D scenes is a multi-hypothesis inference problem, especially in transparent scenarios with multiple plausible depths.\footnote{This work focuses on ambiguous scenes with \textbf{two} visible depth layers, leaving scenes with more than two layers for future work.} To address this, we propose: 1) the \texttt{MD-3k} benchmark, which includes multi-layer spatial relationship labels, 2) new metrics for quantifying single-layer depth ambiguity and evaluating multi-layer spatial relationship accuracy, and 3) a training-free spectral prompting method, \textit{i.e.}, {Laplacian Visual Prompting}, to estimate multi-layer depth.

%propose a framework for bias-corrected multi-layer depth estimation, incorporating a novel benchmark (\texttt{MD-3k}) and evaluation metrics to inform probabilistic inference.\xl{can you discuss in the real order subsequently? such as ..., we propose (1) a benchmark, (2) metrics, (3) new method.}\xh{Good idea! Let me modify it.}

\subsection{\texttt{\textbf{\texttt{MD-3k}}} Benchmark: Quantifying Multi-Layer Spatial Relationships}

The \texttt{MD-3k} benchmark quantifies spatial bias in depth estimation and evaluates multi-layer depth quality in ambiguous scenarios, providing an empirical foundation for assessing layer preference bias in pre-trained models.
%This quantification directly informs the design of our bias-correction prior, as described in Section~\ref{sec:method}.

\vspace{0.5mm}
\noindent \textbf{Benchmark construction.} \texttt{MD-3k} consists of 3,161 RGB images sourced from the GDD dataset~\cite{mei2020don}, selected for depth ambiguity, such as transparency. Following previous spatial relationship benchmarks for non-ambiguous scenes (\textit{e.g.}, \texttt{DIW}~\cite{diw} and \texttt{DA-2k}~\cite{yang2024depth}), we randomly sample a sparse point pair within the ambiguous region for each image. Expert annotators assigned pairwise depth order labels to points both on and behind transparent surfaces, generating two annotation layers. As shown in Fig.~\ref{fig:benchmark}, each sample includes an RGB image, segmentation masks, and two types of spatial relationship labels (\textit{near} and \textit{far}) for point pairs representing multi-layer depths. Annotation accuracy was rigorously validated through multi-round expert review.  \footnote{\texttt{MD-3k} datasheet~\cite{datasheet} is provided in the supplementary material.}

\vspace{0.5mm}
\noindent \textbf{Benchmark statistics.} The full \texttt{MD-3k} dataset, referred to as \textbf{\textit{overall}}, is divided into two subsets with different multi-layer spatial relationships for fine-grained analysis:

\begin{itemize}
    \item \textbf{\textit{Same}} subset (1,783 point pairs): Consistent multi-layer relative depth ordering for each point pair.
    \item \textbf{\textit{Reverse}} subset (1,378 point pairs): Reversed multi-layer relative depth ordering for each point pair.
\end{itemize}
These subsets facilitate the evaluation of depth estimation models under varying conditions of multi-layer spatial ambiguity and relative depth consistency.

Fig.~\ref{fig:benchmark-distribution} summarizes key statistics of ambiguous regions in the \texttt{MD-3k} benchmark. The left panel shows a histogram of the ambiguous-to-total area ratio per sample, capturing diverse ambiguity levels from minimal to near-total scene ambiguity. The  right panel's heatmap indicates a balanced spatial distribution with a slight center bias, resembling a Gaussian pattern that reflects natural scene compositions while minimizing regional biases.

\subsection{Metrics: Revealing Depth Bias and Quantifying Multi-Layer Depth Accuracy}

We propose metrics to capture depth prediction layer bias and quantify the accuracy of multi-layer depth estimation.

\vspace{0.5mm}
\noindent \textbf{Spatial Relationship Accuracy (SRA($i$)).} SRA($i$) measures the fraction of point pairs $\mathcal{P}$ with correct relative depth ordering for each depth layer $i \in \{1, 2\}$:
\vspace{-2mm}\begin{equation} \label{eq:sra_i}
    \begin{split}
        &\text{SRA}(i) = \frac{1}{|\mathcal{P}|} \sum_{(P_1,P_2) \in \mathcal{P}} \\
        & \quad \mathbb{I}\Bigl( \text{sign}(\hat{d}_1^{(i)} - \hat{d}_2^{(i)}) = \text{sign}(d_1^{(i)} - d_2^{(i)}) \Bigr),
    \end{split}
\end{equation}
where $\hat{d}_j^{(i)}$ and $d_j^{(i)}$ represent the predicted and ground truth depths at point $P_j$ for layer $i$, respectively.

\vspace{0.5mm}
\noindent \textbf{Depth Layer Preference ($\alpha(f_\theta)$).} $\alpha(f_\theta)$ quantifies the bias of a depth estimation model $f_\theta$ towards one of the layers for layered scenes in its predictions. It is computed as the difference in SRA between the two layers:
\begin{equation}\label{eq:depth_layer_preference}
\alpha(f_\theta) = \text{SRA}(2) - \text{SRA}(1).
\end{equation}
A positive value, \textit{i.e.}, $\alpha(f_\theta) > 0$, indicates a preference for the second layer, while a negative value, \textit{i.e.}, $\alpha(f_\theta) < 0$, indicates a preference for the first layer. A larger absolute value, \textit{i.e.}, $|\alpha(f_\theta)|$, signifies a stronger bias. %This metric is critical for identifying models that favor one layer over another in ambiguous scenes, such as those with transparency or occlusion.

\vspace{0.5mm}
\noindent \textbf{Multi-Layer Spatial Relationship Accuracy (ML-SRA).} ML-SRA measures the proportion of point pairs where the predicted relative depth ordering is correct in both layers simultaneously, which is formulated as:
\vspace{-1mm}\begin{equation}
\begin{split}
&\text{ML-SRA} = \frac{1}{|\mathcal{P}|} \sum_{(P_1, P_2) \in \mathcal{P}}  \\
&\mathbb{I} \left( \bigwedge_{k=1}^{2}  
\text{sign}(\hat{d}_1^{(k)} - \hat{d}_2^{(k)}) =  
\text{sign}(d_1^{(k)} - d_2^{(k)}) \right).
\end{split}
\end{equation}

\subsection{Laplacian Visual Prompting for Multi-Layer Depth Decoupling}

\begin{figure}[t!]
    \centering
    \setlength{\abovecaptionskip}{0.2cm}
    \includegraphics[width=0.48\textwidth]{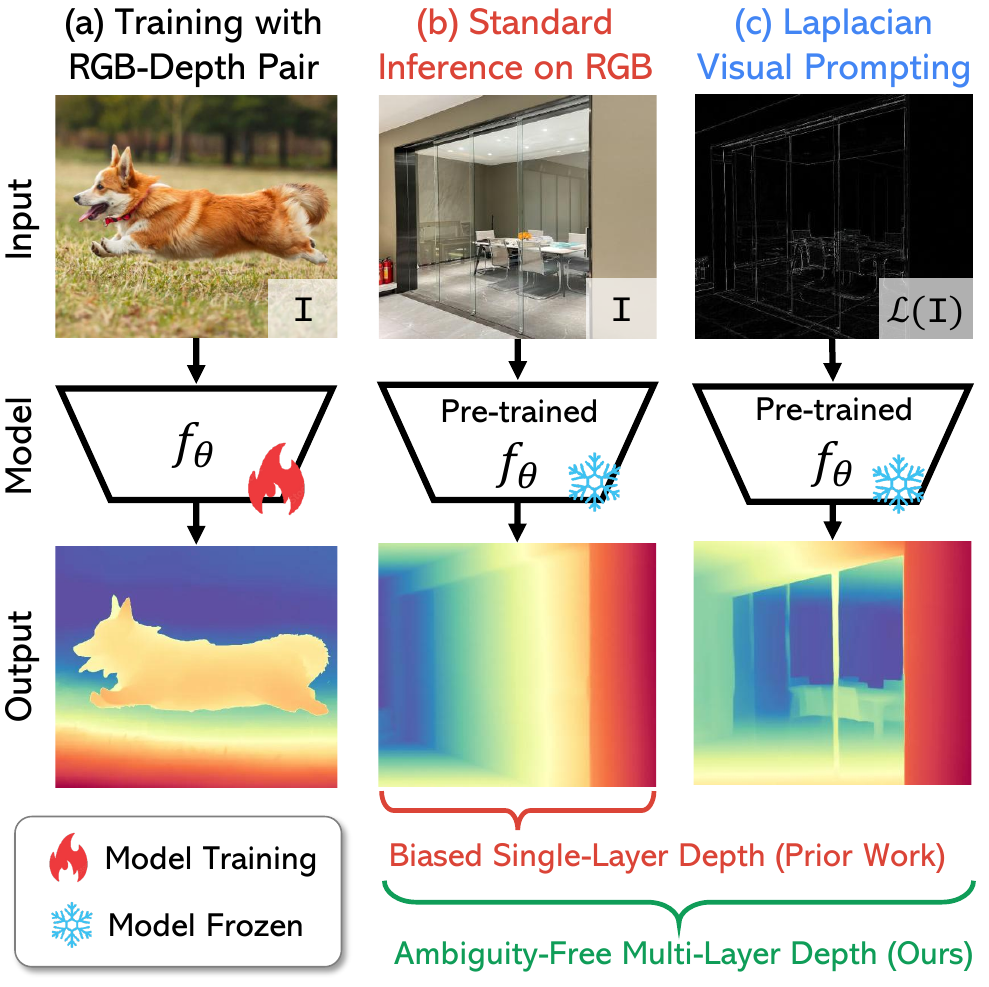}
    \caption{\textbf{Multi-layer depth with Laplacian Visual Prompting (LVP).} (\textbf{a}) Paired RGB-depth training of a domain-specific or domain-agnostic depth estimation model. (\textbf{b}) Standard inference via RGB input: single-layer depth on transparent glass. (\textbf{c}) Model inference via LVP: hidden depth revealing occluded objects, such as tables and chairs, behind the glass.}
    \label{fig:method}\vspace{-1mm}
\end{figure}

As shown in Fig.~\ref{fig:method}, we propose \emph{Laplacian Visual Prompting} (LVP), a visual prompting technique designed to decouple multi-layer depth estimation by leveraging spectral components to resolve depth ambiguities in 3D scenes. LVP does not require retraining the depth model; instead, it employs a pre-trained monocular depth estimator to generate multiple depth hypotheses from a single RGB image. We posit that the latent depth distributions revealed by LVP can enhance depth estimation accuracy in scenarios with inherent depth ambiguity.

\vspace{0.5mm}
\noindent\textbf{Probabilistic modeling of  multi-hypotheses depth.}
To address depth ambiguity in monocular images, we propose a probabilistic model that predicts an {ordered} set of depth hypotheses, $\{\mathcal{D}_1, \mathcal{D}_2\}$, conditioned on the input image $\mathcal{I}$. To capture the relative depth ordering, we introduce a binary latent variable $\mathcal{O} \in \{0,1\}$, where $\mathcal{O}=1$ indicates that $\mathcal{D}_1 \prec \mathcal{D}_2$ (\textit{i.e.}, $\mathcal{D}_1$ is closer than $\mathcal{D}_2$) and $\mathcal{O}=0$ denotes that $\mathcal{D}_2 \prec \mathcal{D}_1$ (\textit{i.e.}, $\mathcal{D}_2$ is closer than $\mathcal{D}_1$).

Rather than marginalizing over all possible orderings, we directly predict the ordered pair $(\mathcal{D}_1, \mathcal{D}_2)$ based on the sampled ordering $\mathcal{O}$:
\begin{equation} \label{eq:joint_z}
p(\mathcal{D}_1, \mathcal{D}_2 \mid \mathcal{I}) = p(\mathcal{D}_1, \mathcal{D}_2 \mid \mathcal{O}, \mathcal{I})\, p(\mathcal{O} \mid \mathcal{I}).
\end{equation}

Assuming that $\mathcal{D}_1$ and $\mathcal{D}_2$ are independently estimated from $\mathcal{I}$ and that the single-layer depth prediction model is agnostic to the ordering $\mathcal{O}$, we derive:
\begin{equation}
p(\mathcal{D}_1, \mathcal{D}_2 \mid \mathcal{O}, \mathcal{I}) \propto p(\mathcal{D}_1 \mid \mathcal{I})\, p(\mathcal{D}_2 \mid \mathcal{I}).%\xl{typo?}
\end{equation}
Substituting into Eq.~\eqref{eq:joint_z} yields:
\begin{equation}
p(\mathcal{D}_1, \mathcal{D}_2 \mid \mathcal{I}) \propto p(\mathcal{D}_1 \mid \mathcal{I})\, p(\mathcal{D}_2 \mid \mathcal{I})\, p(\mathcal{O} \mid \mathcal{I}),
\end{equation}
where $p(\mathcal{D}_1 \mid \mathcal{I})$ and $p(\mathcal{D}_2 \mid \mathcal{I})$ represent the marginal likelihoods of the depth estimates for the two layers, and $p(\mathcal{O} \mid \mathcal{I})$ encodes the probability of the relative depth ordering. The relative ordering can be determined from the sign of layer preference $\alpha(f_\theta)$, as defined in Eq.~\eqref{eq:depth_layer_preference}.

\vspace{0.5mm}
\noindent\textbf{Laplacian transformation for depth disambiguation.}
Monocular depth estimation often struggles with discontinuities and transparent surfaces, which leads to depth ambiguity. To mitigate this bias, we introduce Laplacian Visual Prompting, a spectral prompting strategy that uses the Laplacian operator to enhance the input image by emphasizing high-frequency details like object boundaries and edges. The Laplacian, a second-order derivative, effectively acts as a high-pass filter in the spatial domain. The 2D spatial Laplacian operator is defined as:
\begin{equation}
\Delta = \frac{\partial^2}{\partial x^2} + \frac{\partial^2}{\partial y^2}.
\end{equation}
For an RGB image $\mathcal{I} \in \mathbb{R}^{H \times W \times 3}$, the Laplacian transformation is applied channel-wise:
\begin{equation}
\mathcal{L}(\mathcal{I}) = \begin{pmatrix} \Delta \mathcal{I}_R; \Delta \mathcal{I}_G; \Delta \mathcal{I}_B \end{pmatrix},
\end{equation}
where $\Delta \mathcal{I}_R$, $\Delta \mathcal{I}_G$, and $\Delta \mathcal{I}_B$ are the Laplacian-transformed red, green, and blue channels, respectively.

In discrete form, the Laplacian operator is approximated via a second-order finite difference scheme using a $3 \times 3$ convolution kernel:
\begin{equation}
\mathcal{M}_{\mathcal{L}} = \begin{bmatrix} 
0 & 1 & 0 \\ 
1 & -4 & 1 \\ 
0 & 1 & 0 
\end{bmatrix}.
\end{equation}

%For a single-channel input $\mathcal{I}_c \in \mathbb{R}^{H \times W}$, the Laplacian-transformed output is computed as:
%\begin{equation}
%\mathcal{L}(\mathcal{I})_c = \mathcal{I}_c \ast \mathcal{M}_{\mathcal{L}},
%\end{equation}
%where $\ast$ denotes the 2D convolution.

\begin{figure}[t!]
\setlength{\abovecaptionskip}{0.1cm}
\includegraphics[width=\linewidth, trim=30 0 0 0, clip]{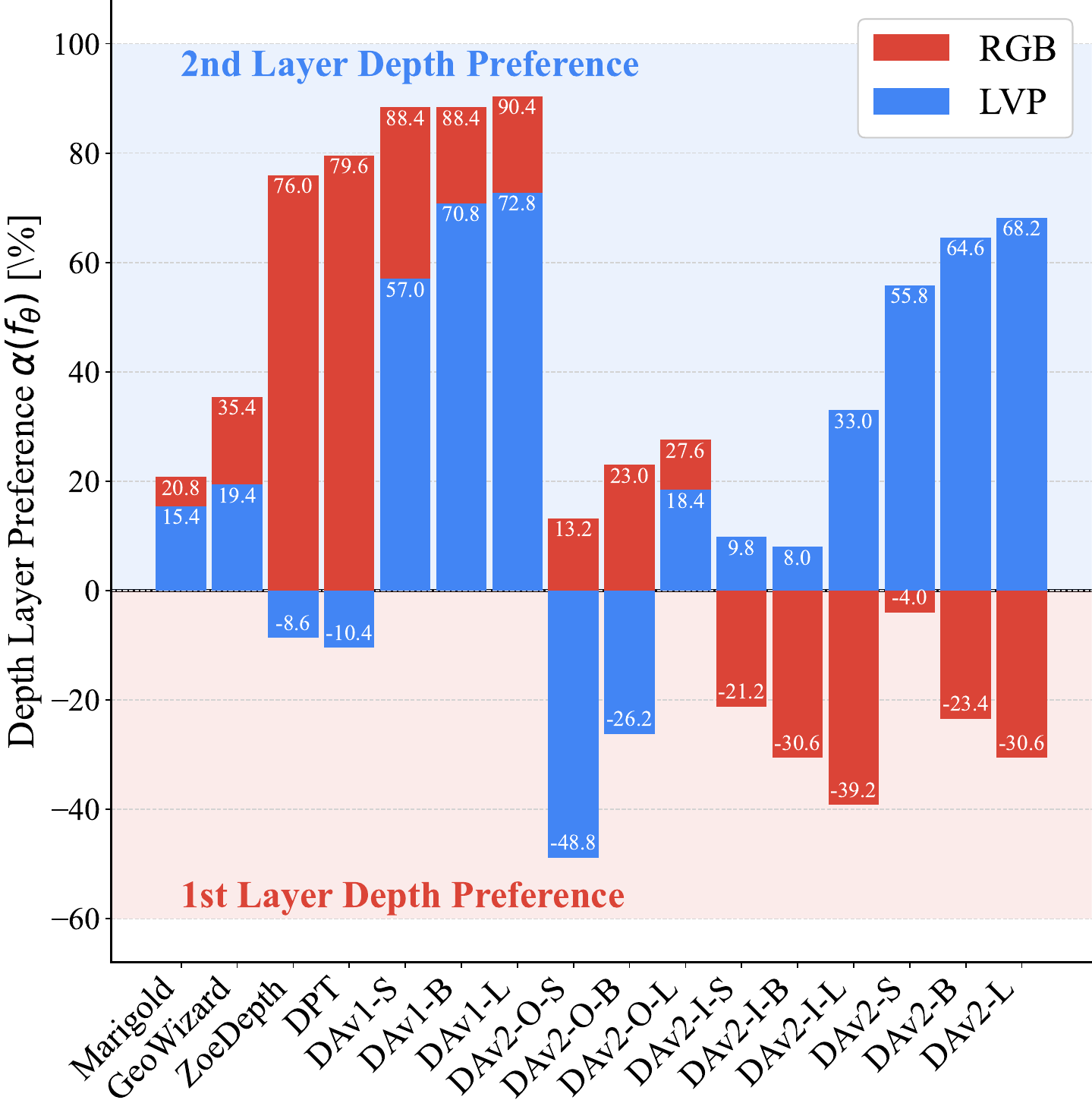}
\caption{\textbf{Depth Layer Preference $\alpha(f_\theta)$ [\%] under RGB and LVP Inputs on \texttt{MD-3k} (\textit{Reverse}).} This figure  highlights that the heterogeneous biases of standard RGB and LVP inputs significantly influence model preference, shifting it between the first and second annotated depth layers for certain models. This demonstrates how input modality can alter depth layer bias.}
    \label{tab:main_table}\vspace{-2mm}
\end{figure}

\noindent\textbf{Multi-hypothesis depth estimation.}
We apply a pre-trained monocular depth model $f_\theta$ to the original RGB  image and its Laplacian-transformed version separately, which generates complementary depth hypotheses:
\begin{equation}
\mathcal{D}_1 = f_{\theta}(\mathcal{I}), \quad \mathcal{D}_2 = f_{\theta}(\mathcal{L}(\mathcal{I})).
\end{equation}
These two independent depth predictions, $\mathcal{D}_1$ and $\mathcal{D}_2$, are combined with the latent ordering probability $p(\mathcal{O} \mid \mathcal{I})$ as described in Eq.~\eqref{eq:joint_z}. This formulation addresses ambiguity by explicitly representing multiple depth hypotheses.

\section{Experiments}

%\subsection{Experimental Overview}
To address the challenges of biased spatial understanding and unlock the potential of multi-hypothesis depth estimation, we raise the following fundamental questions:
 1) \textbf{Depth bias} (Sec.\ref{ssec:single_layer_bias}): In ambiguous scenes, what depth layer biases do existing models exhibit? 
2) \textbf{LVP enhancement} (Sec.\ref{ssec:towards_mh_sfm}):  Can LVP effectively enhance multi-layer spatial understanding? 
3) \textbf{Scaling laws} 
 (Sec.\ref{ssec:scaling_laws}): How does model scale influence LVP-enhanced spatial understanding? 
4) \textbf{Practical benefits} 
 (Sec.\ref{ssec:practical_benefits}): What are the practical advantages of LVP-driven multi-hypothesis depth?
5) \textbf{LVP design} (Sec.\ref{ssec:ablation_study}): How do LVP's design choices impact performance under ambiguity?

% where S, B, and L denote the small, base, and large variants with different model backbone sizes
\vspace{0.5mm}
\noindent\textbf{Baseline models.} We assess bias and evaluate the effectiveness of our proposed multi-hypothesis depth estimation method via LVP across diverse baseline models, including Depth Anything V1/V2 (DAv1/2-{S,B,L} with ViT backbones~\cite{depth_anything,yang2024depth}). These models include general (DAv1/2), indoor (DAv2-I), and outdoor (DAv2-O) fine-tuned variants. Additional models include the discriminative models DPT~\cite{dpt} and ZoeDepth~\cite{zoedepth}, as well as the generative models Marigold~\cite{marigold} and GeoWizard~\cite{geowizard}.

\subsection{Probing Single-Layer Depth Prediction Bias}
\label{ssec:single_layer_bias}

We first analyze depth layer preference bias, which is defined in Eq. (\ref{eq:depth_layer_preference}), for baseline models, revealing inherent biases in predicting closer or farther surfaces in ambiguous regions using standard RGB  input. We then explore whether LVP can alter these biases by introducing complementary depth hypotheses to enrich RGB predictions.

\vspace{0.5mm}
\noindent\textbf{Heterogeneous depth layer bias under standard RGB input.} In Fig.~\ref{tab:main_table}, we observe significant heterogeneity in depth layer prediction preferences across models. Some models (\textit{e.g.}, DAv2, DAV2-I) exhibit a bias towards the first depth layer, \textit{i.e.}, $\alpha(f_\theta)<0$, while others favor the second depth layer. In addition, models with the same architecture fine-tuned on different datasets (indoor/outdoor) can exhibit opposing depth biases, \textit{e.g.}, DAv2-I and DAv2-O. This highlights how training data can hardwire assumptions about scene structure. %, limiting generalization to diverse ambiguous scenarios.

\vspace{0.5mm}
\noindent\textbf{LVP modulates depth prediction preference.} Comparing RGB and LVP results in Fig.~\ref{tab:main_table} reveals that LVP effectively \textbf{\textit{reverses}} or \textbf{\textit{attenuates}} depth preferences across all baseline models.  The pronounced impact of LVP on depth preference modification, particularly in models like DAv2, suggests its ability to unlock latent representations. This reveals previously suppressed depth layers and fundamentally reshapes the model's depth interpretation.

\begin{figure}[t!]
\setlength{\abovecaptionskip}{0.2cm}
\includegraphics[width=\linewidth, trim=0 0 25 0, clip]{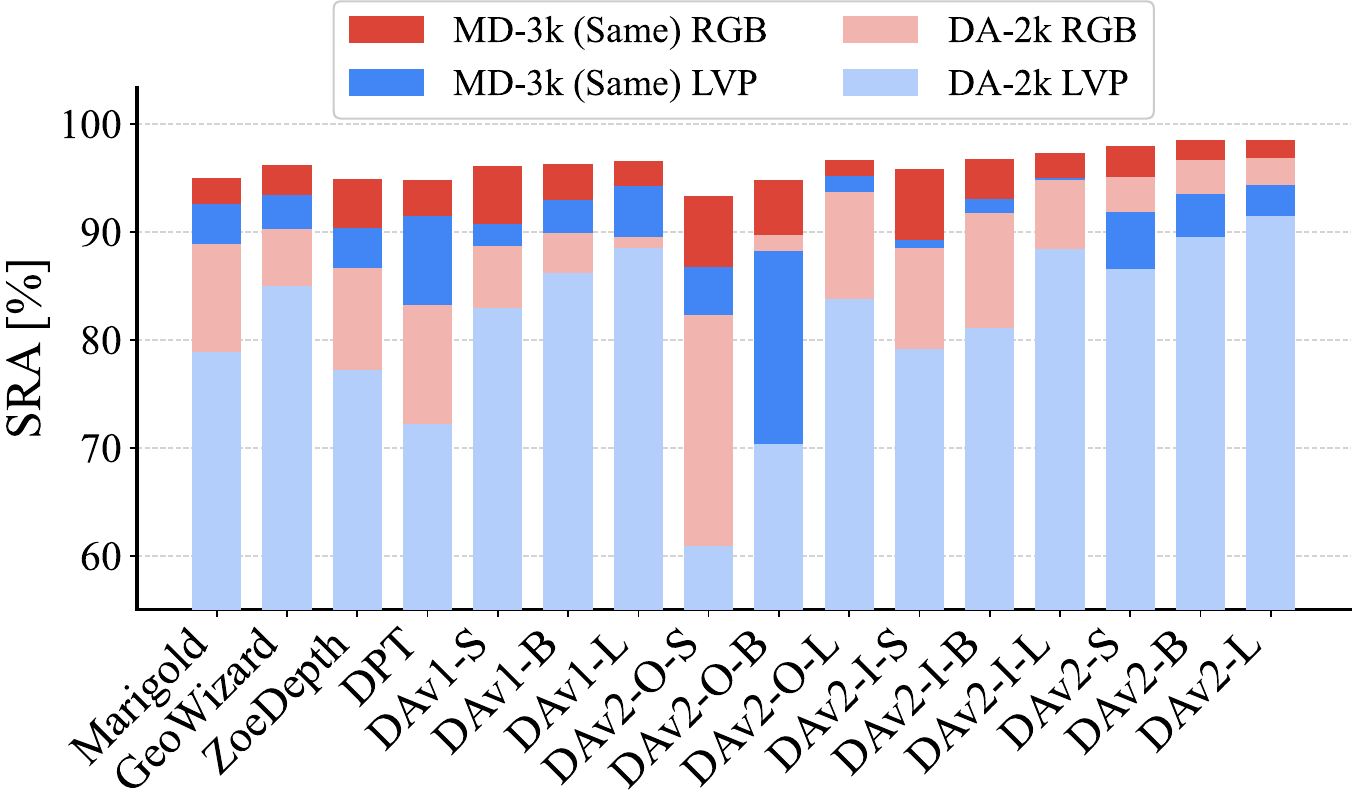}
\caption{\textbf{High Spatial Relationship Accuracy (SRA) [\%] under RGB and LVP inputs}\textbf{ on  the {\textit{same}} subset of  \texttt{MD-3k} and \texttt{DA-2k}~\cite{yang2024depth}} (non-ambiguous benchmark for reference).} 
    \label{fig:ml_sra_same}\vspace{-2mm}
\end{figure}

\vspace{0.5mm}
\noindent\textbf{True depth ambiguity exposes depth biases and remains challenging to tackle.} True depth ambiguity, exemplified by scenes with \textbf{\textit{reverse}} multi-layer spatial relationships (see Fig.~\ref{tab:main_table}), critically reveals depth layer biases. In these ambiguous scenarios, performance becomes inconsistent (\textit{i.e.}, large $|\alpha (f_\theta)|$) across RGB and LVP inputs, highlighting the difficulty in resolving conflicting spatial cues. Conversely, in non-ambiguous scenes with \textbf{\textit{same}} multi-layer relationships (see Fig.~\ref{fig:ml_sra_same}), models achieve consistently high performance, exceeding 85\% SRA under RGB input. This robustness is further supported by the small performance gap between RGB and LVP inputs on the non-ambiguous \texttt{DA-2k} benchmark, and the comparable RGB performance observed between \texttt{MD-3k} (\textit{same}) and \texttt{DA-2k}. Thus, ambiguous scenes with \textbf{\textit{reverse}} multi-layer relationships serve as a crucial diagnostic tool, effectively exposing the inherent depth biases and limitations of current depth baseline models.

\subsection{Multi-Layer Spatial Relationship Accuracy}
\label{ssec:towards_mh_sfm}

\begin{figure}[t!]
\setlength{\abovecaptionskip}{0.2cm}
\includegraphics[width=\linewidth, trim=0 0 25 0, clip]{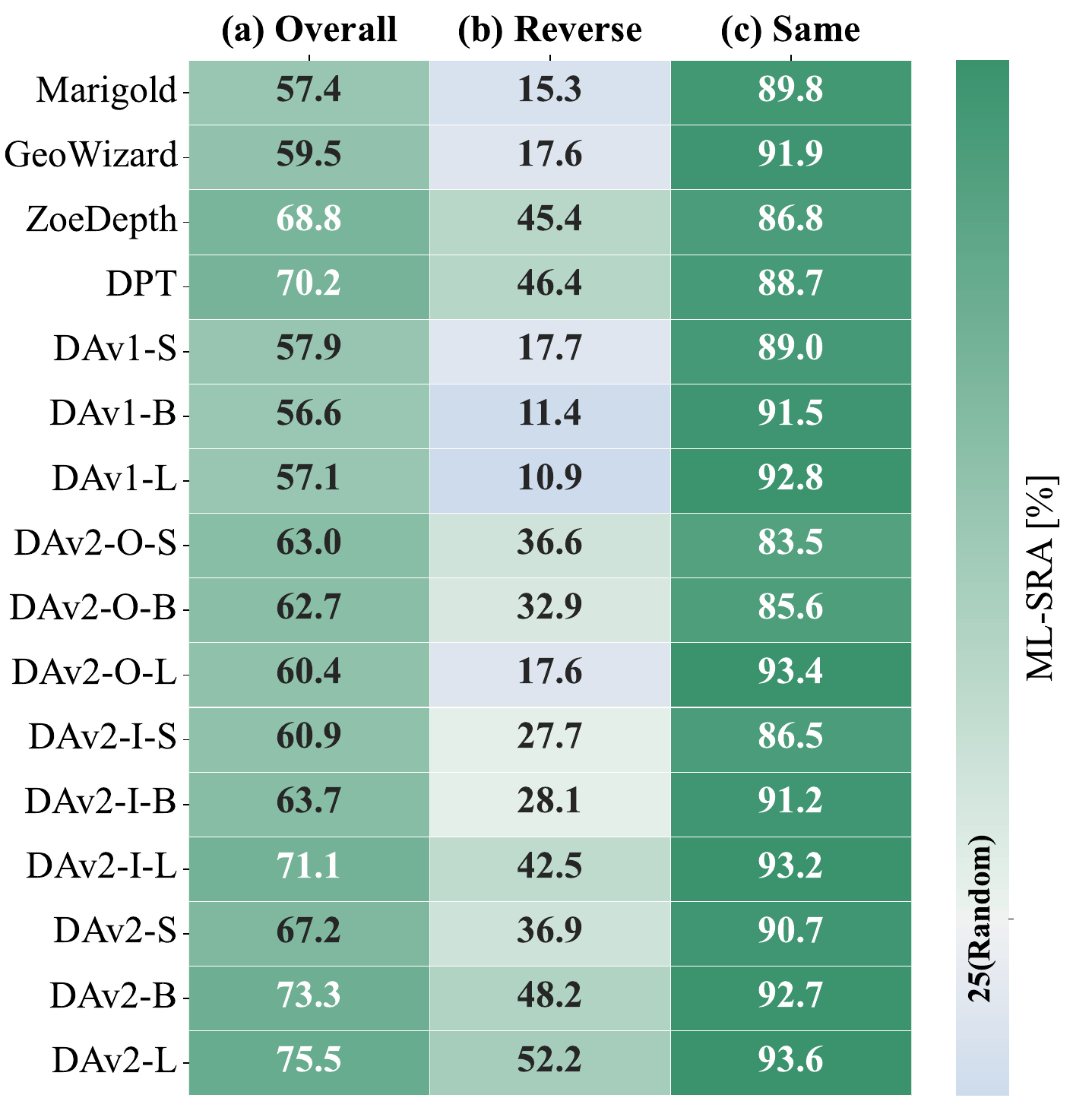}
\caption{\textbf{Multi-Layer Spatial Relationship Accuracy (ML-SRA) [\%] of our LVP-empowered multi-layer depth on \texttt{MD-3k}}. Effective performance gains of LVP-derived multi-depth  over random guess (25\%) are highlighted in \textcolor[HTML]{0F9D58}{green boxes}. }
    \label{tab:merged_ml_sra}\vspace{-2mm}
\end{figure}

Fig.~\ref{tab:merged_ml_sra} shows the Multi-Layer Spatial Relationship Accuracy (ML-SRA) achieved by our multi-hypothesis depth estimation method, which combines depth estimates from both RGB images and LVP inputs. This evaluation demonstrates the effectiveness of Laplacian Visual Prompting in generating complementary depth hypotheses beyond RGB-based inference, enabling ambiguity-free spatial understanding across diverse baselines. 

\vspace{0.5mm}
\noindent\textbf{Latent multi-layer knowledge suggests potential for MH-SFMs, unlocked by LVP.} Despite being trained on single-layer depth data, some models implicitly capture multi-layer spatial relations. For example, DAv2, ZoeDepth, and DPT, when prompted with LVP, achieve non-trivial ML-SRA scores in challenging reverse spatial relationships (see Fig.~\ref{tab:merged_ml_sra}b). This demonstrates that LVP effectively elicits this latent knowledge, suggesting that these models have the potential to be adapted into MH-SFMs, capable of representing and reasoning about multiple depth hypotheses. The fact that LVP is able to unlock this hidden potential highlights its significance as a key enabler for multi-hypothesis depth estimation.

\vspace{0.5mm}
\noindent\textbf{Challenges in }\textit{\textbf{reverse}}\textbf{ relationships highlight the need for explicit ambiguity modeling, even with LVP.} Accurately resolving depth ambiguity in \textit{reverse} multi-layer spatial relationships remains challenging, even when using LVP to generate multi-layer depth estimates. The performance gap between \textit{same} and \textit{reverse} relationships highlights the difficulty of handling conflicting spatial cues and the limitations of relying solely on implicit priors, even when augmented by LVP. The reduced ML-SRA of domain-finetuned DAv2 models further suggests that optimizing for single-domain performance can hinder generalization to multi-layer scenes, reinforcing the need for models that can explicitly model and resolve ambiguity, rather than relying on domain-specific heuristics, even when combined with LVP-based prompting.

\begin{figure}[t!]
\setlength{\abovecaptionskip}{0.1cm}
     \includegraphics[width=0.48\textwidth]{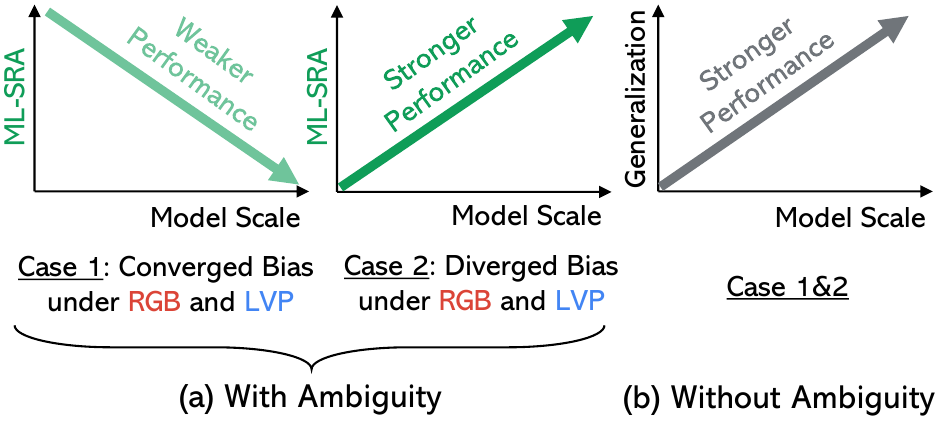}
\caption{\textbf{Scaling laws of spatial understanding: performance trends in ambiguous vs. non-ambiguous scenes.}  (\textbf{a}) In ambiguous scenes, converged depth bias (Case 1) leads to weaker performance with scale, while diverged bias (Case 2) yields stronger performance. (\textbf{b}) In non-ambiguous scenes, performance consistently improves with model scale, showing stronger generalization to the \textit{out-of-distribution} LVP input.}
    \label{fig:summary_scaling_law}\vspace{-2mm}
\end{figure}

\subsection{Scaling Laws of Spatial Understanding}
\label{ssec:scaling_laws}
%Developing generalist foundation models requires understanding performance scaling~\cite{fan2024scaling,bai2024sequential}. We investigate how model scale impacts depth layer bias, multi-layer depth estimation in ambiguous scenes, and single-layer depth estimation in non-ambiguous scenes using Laplacian Visual Prompting (LVP), providing insights for building more reliable large-scale spatial foundation models.

 %To provide a high-level overview of these scaling behaviors, Fig.~\ref{fig:summary_scaling_law} summarizes the key performance trends observed in both ambiguous and non-ambiguous scenes as model scale increases.  As depicted, \textbf{the impact of model scaling on spatial understanding is nuanced and context-dependent}.  Specifically, in ambiguous scenes, we observe divergent performance scaling depending on whether the model exhibits converged or diverged depth bias under RGB and LVP inputs.  Conversely, in non-ambiguous scenes, a more consistent pattern of performance improvement with scale emerges. % The following subsections delve into the detailed empirical evidence supporting these summarized trends, starting with an analysis of depth layer preference bias in ambiguous scenes under RGB input.

Developing generalist foundation models requires understanding performance scaling~\cite{fan2024scaling,bai2024sequential}. We investigate how model scale impacts depth layer bias, multi-layer depth estimation in ambiguous scenes, and single-layer depth estimation in non-ambiguous scenes using Laplacian Visual Prompting (LVP), providing insights for building more reliable large-scale spatial foundation models.

 To provide a high-level overview of these scaling behaviors, Fig.~\ref{fig:summary_scaling_law} summarizes the key performance trends observed in both ambiguous and non-ambiguous scenes as model scale increases.  As depicted, \textbf{the impact of model scaling on spatial understanding is nuanced and context-dependent}, \textbf{\textit{echoing a concurrent work on multi-modal alignment}}~\cite{Tjandrasuwita2025Understanding}. Specifically, in ambiguous scenes, we observe divergent performance scaling depending on whether the model exhibits \textbf{\textit{converged}} or \textbf{\textit{diverged}} depth bias under RGB and LVP inputs.  This divergence suggests that in scenarios with high spatial ambiguity (akin to high \textit{uniqueness} in multi-modal data~\cite{Tjandrasuwita2025Understanding}), simply increasing model scale does not uniformly translate to improved performance. Instead, the \textit{nature} of representation learning, specifically the depth bias, becomes a critical factor. Conversely, in non-ambiguous scenes, a more consistent pattern of generalization improvement with scale emerges, aligning with the expected benefits of larger model capacity in less challenging scenarios where redundancy is higher and ambiguity is lower.

\begin{figure}[t!]
    \centering \setlength{\abovecaptionskip}{0.2cm}
      \includegraphics[width=0.48\textwidth]{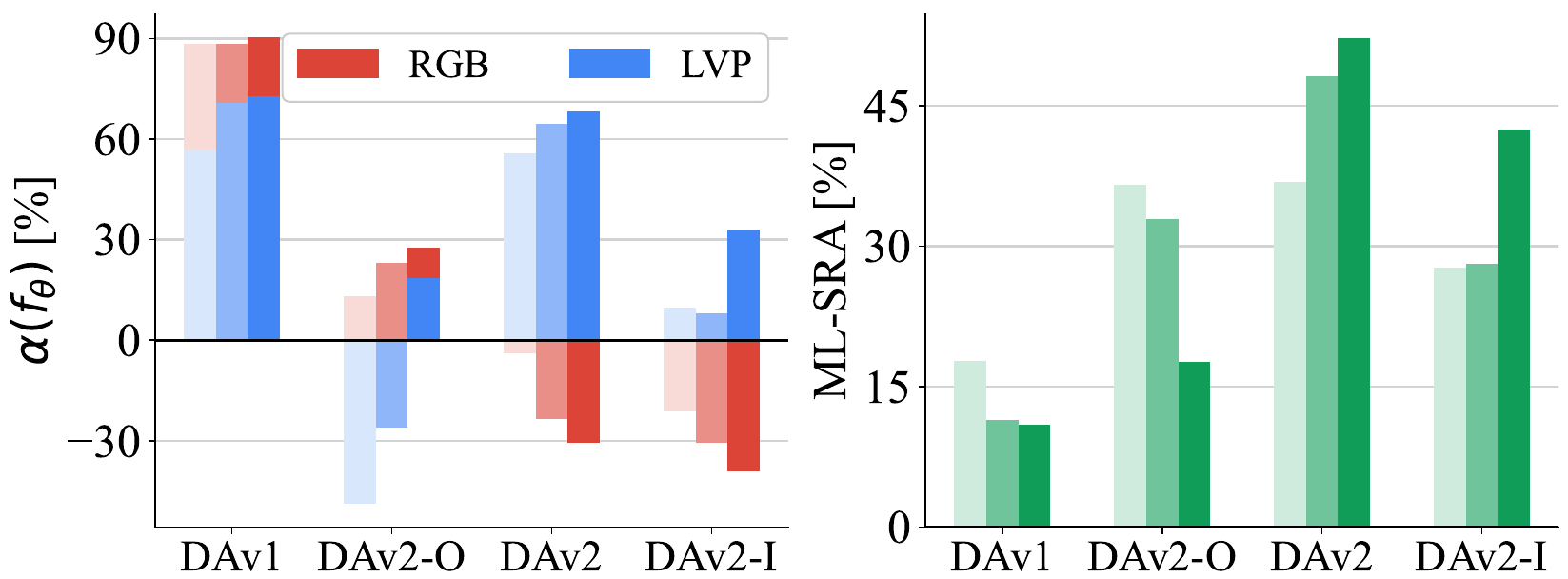}
\caption{\textbf{Performance in ambiguous scenes (\texttt{MD-3k} \textit{reverse} subset) as model scale increases.}  \textbf{Left:} Depth Layer Preference $\alpha(f_\theta)$ [\%]. \textbf{Right:} Multi-Layer Spatial Relationship Accuracy (ML-SRA) [\%]. Bars within each group represent small, base, and large model variants (left to right).}
    \label{fig:scale_ambiguous}
    \vspace{1mm}

\setlength{\abovecaptionskip}{0.2cm}
    \includegraphics[width=0.48\textwidth]{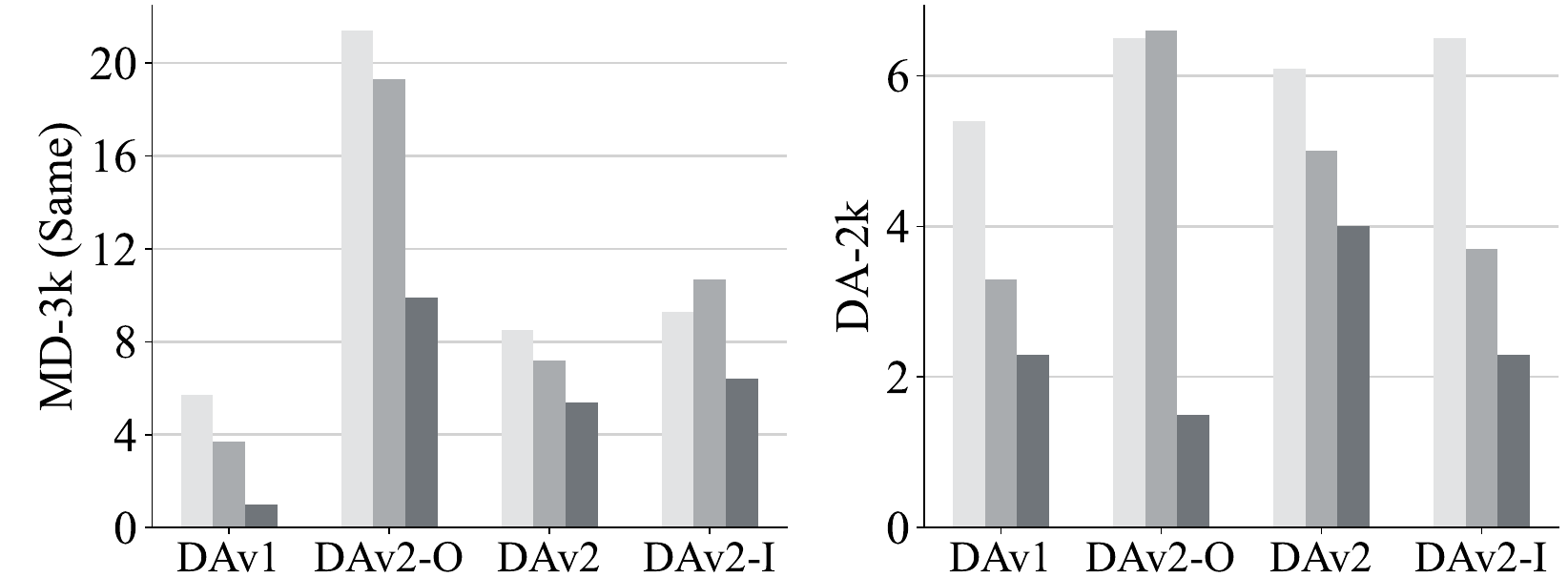}
\caption{\textbf{Generalization to LVP input in non-ambiguous scenes as model scale increases}, measured by the SRA [\%] gap between RGB and LVP inputs on the \textit{same} subset of  \texttt{MD-3k} (\textbf{Left}) and the non-ambiguous benchmark \texttt{DA-2k} (\textbf{Right}). }
\label{fig:scale_nonambiguous}\vspace{-4mm}
\end{figure}

\vspace{0.5mm}
\noindent\textbf{Model scale amplifies depth layer preference bias under RGB input.} As shown in the left panel of Fig.~\ref{fig:scale_ambiguous}, larger models tend to exhibit a stronger preference for certain depth layers under RGB input in ambiguous scenes with \textbf{\textit{reverse}} multi-layer spatial relationships. DAv1 and DAv2-O models demonstrate a growing preference for the second depth layer while  DAv2 and DAv2-I models demonstrating an increasing preference for the first layer.

\vspace{0.5mm} \noindent\textbf{Divergent depth bias elicit stronger multi-layer depth prediction  with scale.} As shown in the right panel of Fig.~\ref{fig:scale_ambiguous}, larger models can exhibit a stronger divergence in depth layer preference based on input modality (RGB vs. LVP) in ambiguous scenes. Some models (DAv1, DAv2-O) \textbf{\textit{converge}} towards a consistent second-layer preference, reducing multi-layer accuracy. Others (DAv2, DAv2-I) show a \textbf{\textit{divergence}}, improving ML-SRA with larger model, suggesting more diverse latent representations  that encode multiple depth hypotheses.

\vspace{0.5mm}
\noindent\textbf{Enhanced generalization with increasing model scale in non-ambiguous scenes.}  Fig.~\ref{fig:scale_nonambiguous} shows the performance gap between RGB and LVP inputs narrows as model scales.  This improved generalization is observed in both multi-layer scenes with \textit{same} spatial relationships of \texttt{MD-3k} and non-ambiguous scenes of \texttt{DA-2k}.

\vspace{0.5mm}
\noindent\textbf{Implications for \textit{Generalist Spatial Foundation Models}.} Scaling model size alone is insufficient for MH-SFMs. A nuanced approach, considering model capacity, scene ambiguity, and depth bias, is essential. For ambiguous scenes, representational diversity and bias mitigation may be more effective than simply increasing parameters, while scaling is beneficial for less ambiguous tasks.

\begin{figure}[t!]% \vspace{-1mm}
    \centering \setlength{\abovecaptionskip}{0.2cm}
    \includegraphics[width=0.48\textwidth]{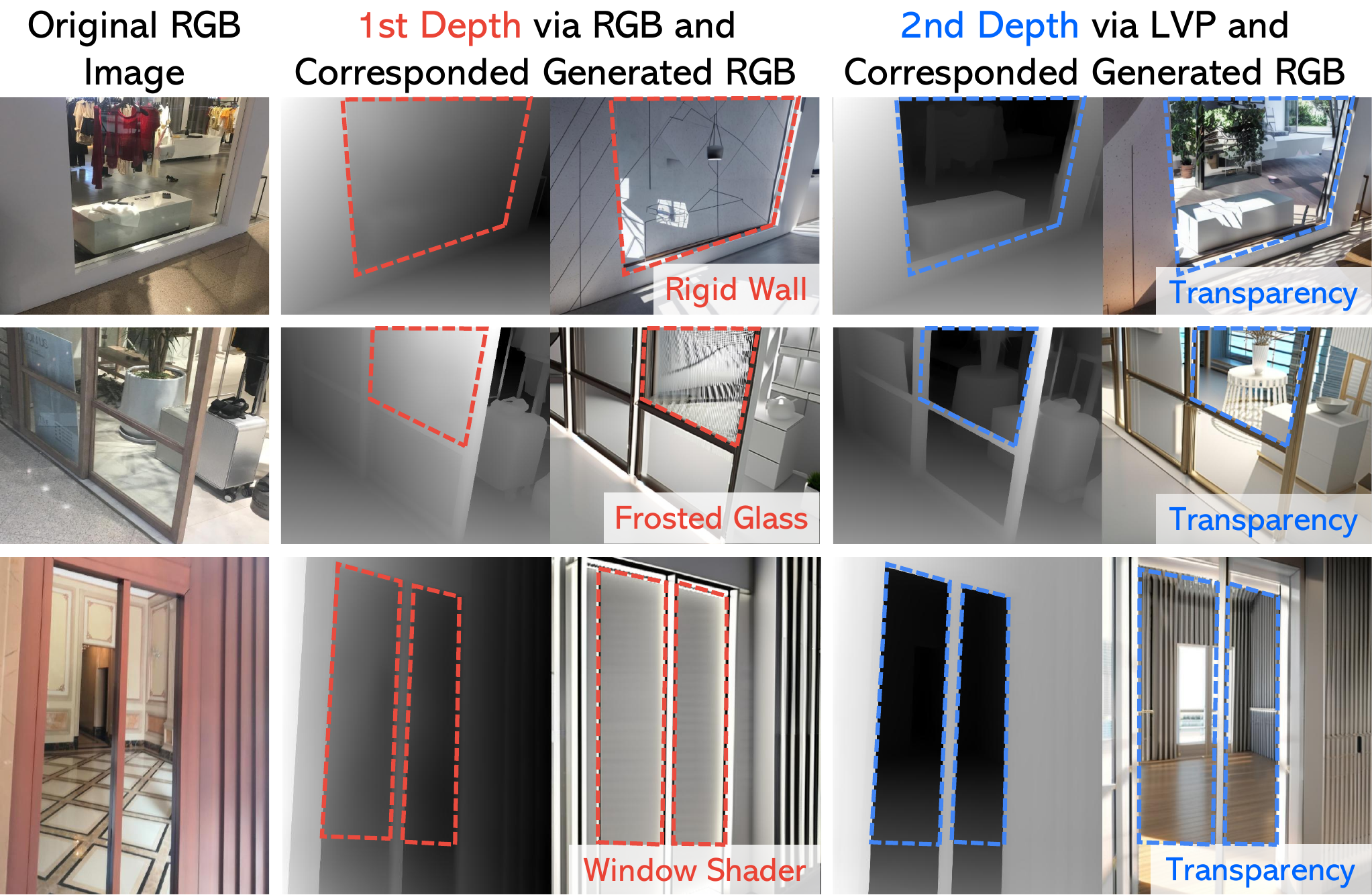}%{figs/depth_to_rgb_cases.pdf}
\caption{\textbf{Flexible 3D-conditioned visual generation. }}
    \label{fig:depth2rgb}\vspace{-2mm}

    \end{figure}
    \begin{figure}[t] \vspace{-0.5mm}
    \centering\setlength{\abovecaptionskip}{0.2cm}
    \includegraphics[width=0.48\textwidth]{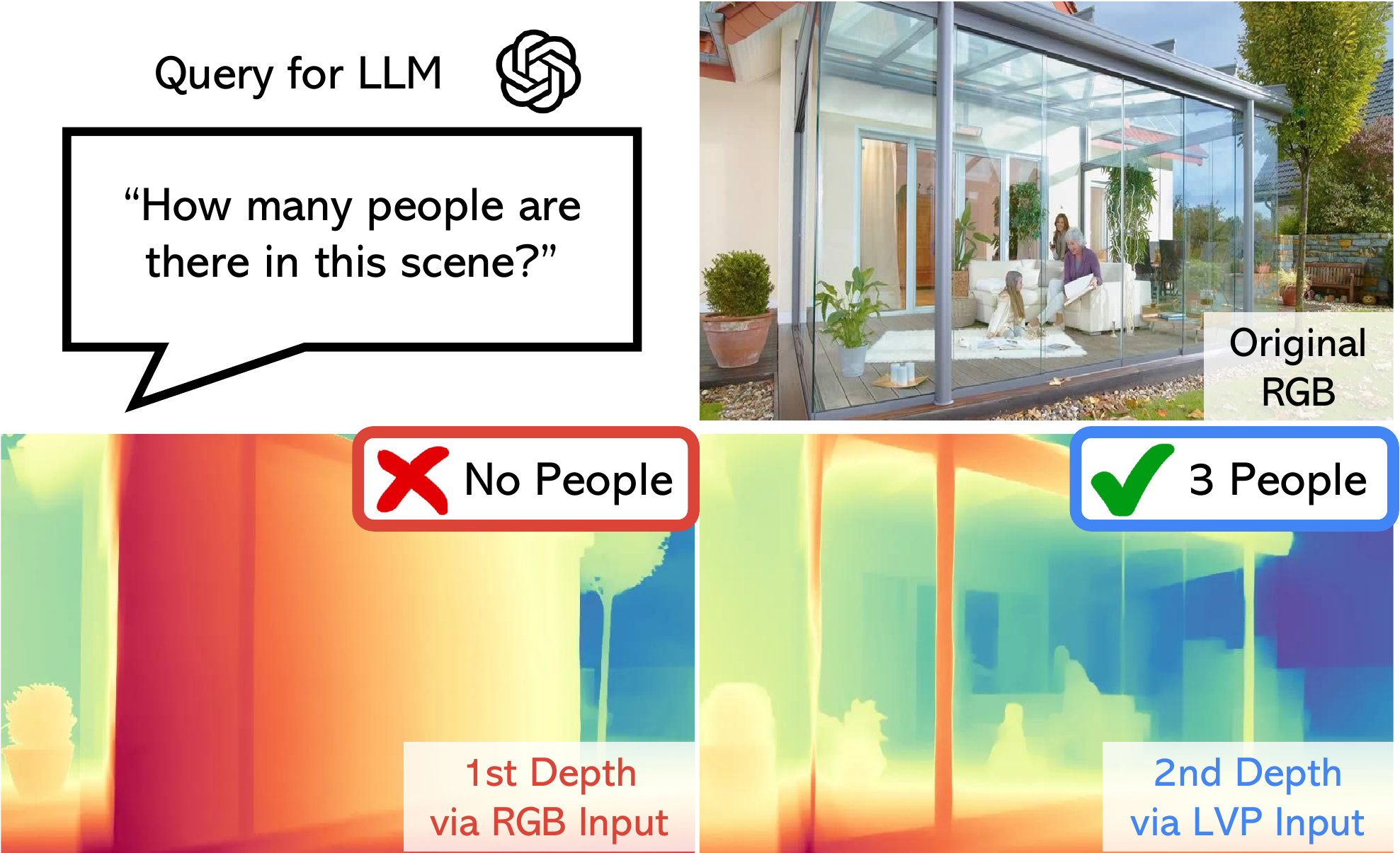}
\caption{\textcolor{black}{\textbf{Robust 3D spatial reasoning with LLM.}}}
    \label{fig:reasoning_llm}\vspace{-3mm}
\end{figure}

\subsection{Applications of Multi-Hypothesis Depth}\label{ssec:practical_benefits}
\begin{figure}
    \centering \setlength{\abovecaptionskip}{0.2cm}
    \includegraphics[width=0.48\textwidth]{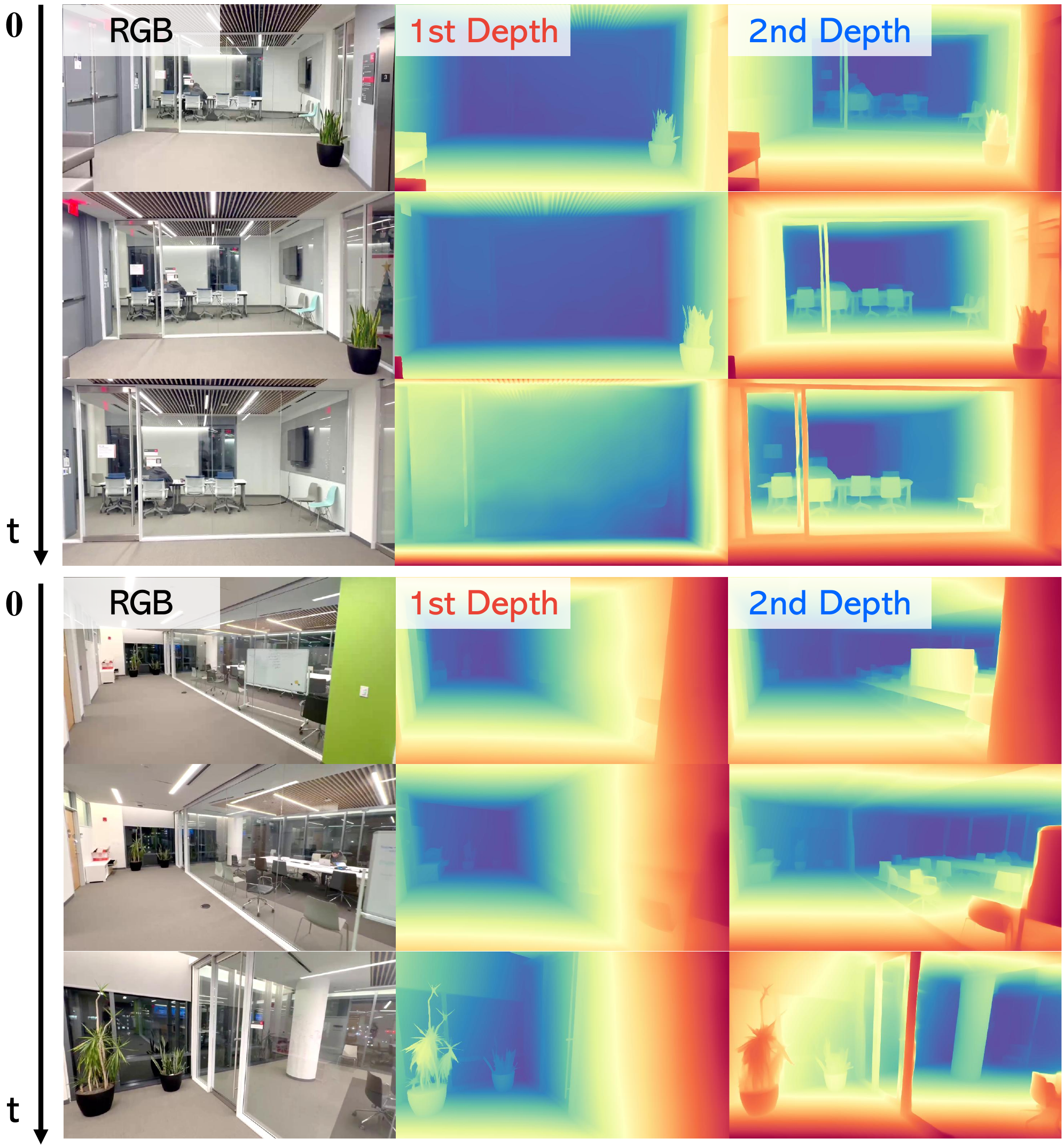}
\caption{\textbf{Consistent multi-layer depth estimation on video.}}
    \label{fig:video_depth}\vspace{-2mm}

\end{figure}

The multi-hypothesis depth predictions enabled by LVP enhance 3D-conditional image generation for ambiguous scenes. This capability supports the creation of complex environments, such as those featuring both transparent and opaque objects (\textit{e.g.}, glass doors and windows), using geometry-conditioned ControlNet~\cite{controlnet} (see Fig.\ref{fig:depth2rgb}). In addition, LVP boosts 3D spatial reasoning through a Multi-modal Large Language Model (LLM), exemplified by precise 3D-grounded human counting with the ChatGPT o3-mini model (see Fig.\ref{fig:reasoning_llm}).  Furthermore, the multi-layer depth estimation underlying these predictions also demonstrates robust consistency when applied to real-world complex videos, as evidenced in Fig.\ref{fig:video_depth}. \footnote{For implementation details and additional qualitative results (including the video demo), please refer to our supplementary material.}

\begin{figure}[t!]
\setlength{\abovecaptionskip}{0.2cm}
     \includegraphics[width=0.48\textwidth]{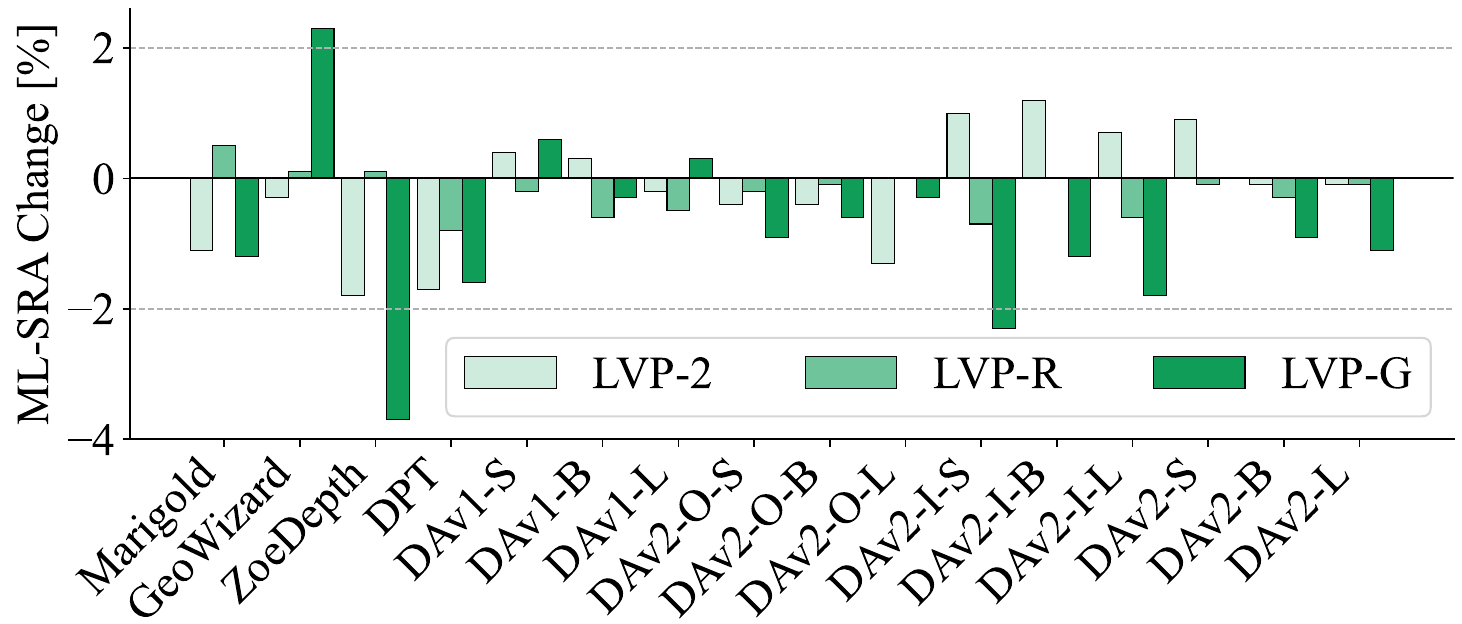}
\caption{\textbf{Ablation study of LVP design}. Overall  ML-SRA [\%] change relative to the default LVP  on \texttt{MD-3k} is shown. }
    \label{fig:ablation_LVP}\vspace{-4mm}
\end{figure}

\subsection{Ablation Study of LVP Design}%: Prompt Engineering and Spectral Sensitivity
\label{ssec:ablation_study} % Added labels for cross-referencing
Fig.~\ref{fig:ablation_LVP} shows that ML-SRA performance is largely unaffected by Laplacian discretization (4-neighbor vs. 8-neighbor in LVP and LVP-2) and kernel sign (LVP-R with reversed convolution vs. LVP), with variations generally within $\pm 3\%$. While grayscale LVP (LVP-G) slightly reduces SRA compared to RGB LVP, the difference is minimal. These results highlight the crucial role of high-frequency information in 3D spatial decoupling.

\section{Conclusion}
\label{sec:conclusion}
We redefine domain-agnostic monocular spatial foundation models as inherently ambiguous, multi-hypothesis problems. To advance this, we introduce Laplacian Visual Prompting (LVP), a training-free technique for multi-layer depth estimation, and \texttt{MD-3k}, the first benchmark for evaluating multi-layer depth under ambiguity. Our analysis highlights significant biases in existing models, revealing the limitations of single-depth estimation. Experiments show that LVP modulates depth biases, enables comprehensive multi-layer estimation, and enhances downstream task robustness and flexibility.

\vspace{0.5mm}
\noindent\textbf{Limitation \& future work.} Future work should explore spatial understanding with a wider range of multi-modal visual prompts, such as learned spectral transformations. While our \texttt{MD-3k} benchmark leverages real-world images containing inherent noise, it is important to evaluate the robustness of LVP under various noise and artifact conditions. Moreover, extending our benchmark to incorporate real-world multi-layer depth annotations—potentially through the integration of multiple physical sensors—would provide a more comprehensive metric for assessing multi-layer depth performance, though this remains a challenging task. Finally, to enable reliable spatial foundation models, future research could explore how to address and characterize more diverse spatial ambiguity, \textit{e.g.}, reflection effects.

\clearpage
\maketitlesupplementary
\setcounter{table}{0}
\setcounter{figure}{0}
\setcounter{section}{0}
\renewcommand{\thetable}{\Alph{table}}
\renewcommand{\thefigure}{\Alph{figure}}
\renewcommand{\thesection}{\Alph{section}}

\section{More Qualitative Results}

\begin{figure}[t!] %\vspace{-5mm}
    \centering\setlength{\abovecaptionskip}{0.2cm}
    \includegraphics[width=0.48\textwidth]{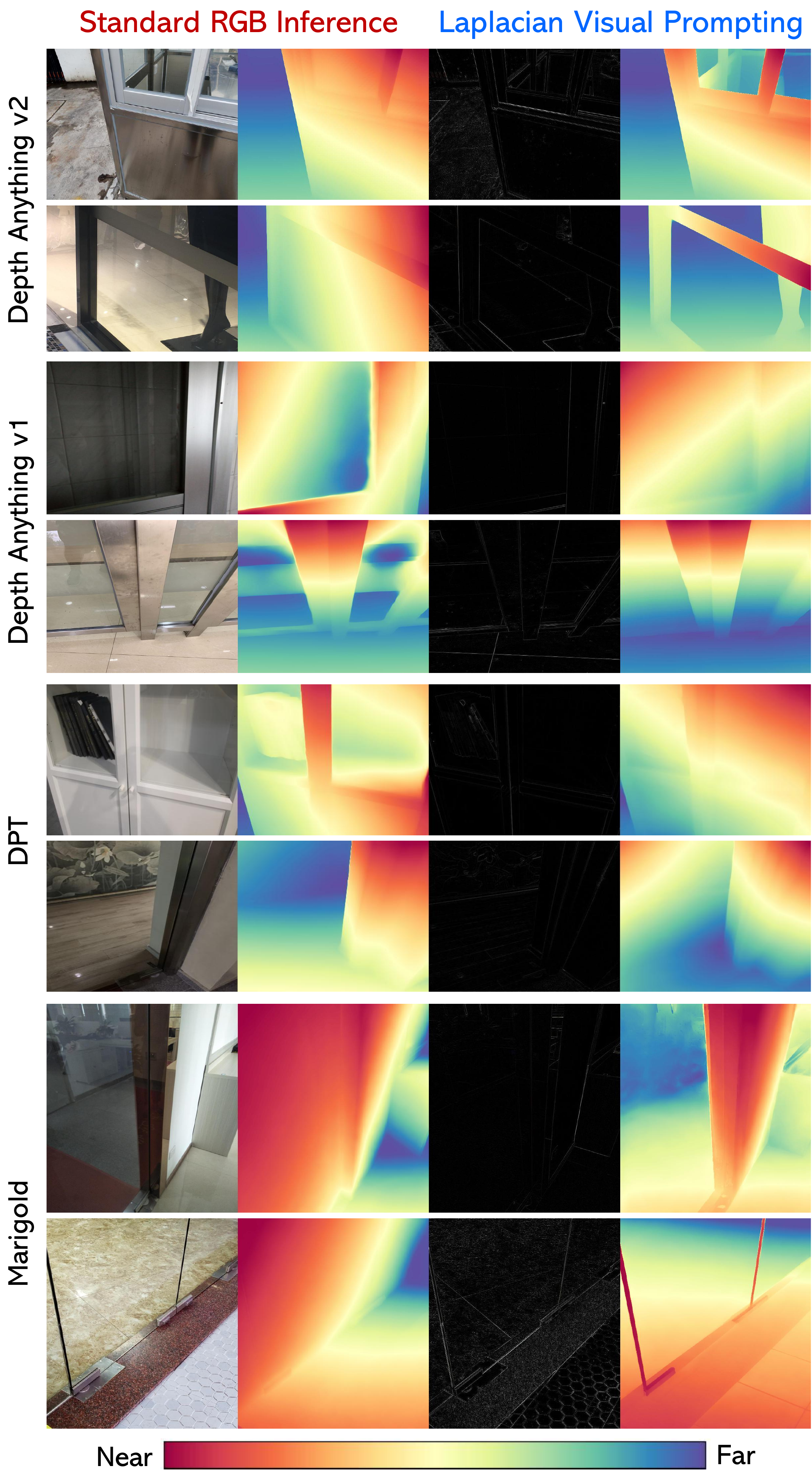}
\caption{\textbf{Unlocking \textit{hidden depth} with Laplacian Visual Prompting on diverse models~\cite{depth_anything,yang2024depth,marigold,dpt}.} Each case shows the RGB input, the estimated depth via RGB, the Laplacian input, and the estimated \textit{depth} via Laplacian input. }
    \label{fig:hidden-depth-appendix}\vspace{-2mm}
\end{figure}

\noindent\textbf{Multi-layer depth decoupling with Laplacian Visual Prompting.} In addition to demonstrating the effectiveness of LVP in revealing hidden depth layers in various models (Fig.~\ref{fig:hidden-depth-appendix}), we further demonstrate its capabilities using the best-performing baseline model, \textit{i.e.}, the Depth Anything v2-Large (DAv2-L) model, in Figures~\ref{fig:hidden_depth_2_1} through~\ref{fig:hidden_depth_2_6}. LVP effectively elicits alternative depth hypotheses, particularly in scenes with transparency and occlusion. Although conventional depth estimations often fail to capture the layered nature of these scenes, collapsing multiple depths into a single layer, Laplacian-prompted depth maps reveal previously hidden depth layers, clearly delineating  transparent surfaces and occluded objects behind the transparent surfaces.

\noindent\textbf{Failure cases in multi-layer depth decoupling with Laplacian Visual Prompting.} Despite significant improvements in multi-layer depth estimation, Laplacian Visual Prompting (LVP) is not immune to failure cases, especially due to its training-free nature. We present these failures in Fig.~\ref{fig:failure_cases}. One type occurs when the initial single-layer depth prediction from the RGB input is already incorrect. In these instances, LVP struggles to correct the depth bias, resulting in inaccurate multi-layer depth predictions. This is particularly problematic when the RGB model’s depth map contains errors in ambiguous regions, limiting LVP’s ability to produce accurate alternative depth hypotheses. The second type of failure happens when LVP predicts a depth layer similar to the RGB output, failing to decouple distinct depth layers. This occurs when LVP cannot extract sufficient high-frequency information from the Laplacian mask to differentiate between near and far surfaces, especially in scenes with low depth contrast or complex occlusions. These cases highlight the challenges in multi-layer depth estimation, underscoring the need for improved baseline foundation models and refined prompts.

\vspace{1mm}
\noindent\textbf{Additional \texttt{MD-3k} benchmark samples.} Fig.~\ref{fig:md3k_case1} presents additional examples from the \texttt{MD-3k} benchmark, our newly introduced data set to evaluate the understanding of multi-layer spatial relationships. These examples highlight the diverse and challenging scenarios within \texttt{MD-3k}, including varying levels of depth ambiguity and transparency. By providing a broader range of scenes, we aim to assess how well models can disambiguate depth layers in multi-layered environments, particularly in real-world images that reflect the complexities and nuances of natural scenes.

\begin{figure}[t!] %\vspace{-5mm}
    \centering \setlength{\abovecaptionskip}{0.2cm}
    \includegraphics[width=0.48\textwidth]{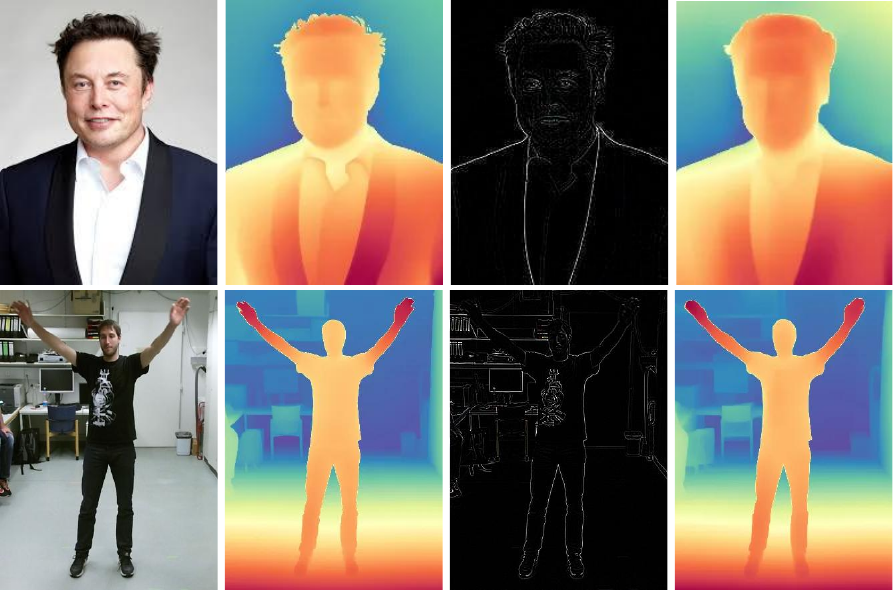}
\caption{\textbf{Privacy-preserving potential of LVP}. Depth estimation results on non-ambiguous scenes using LVP are comparable to those using original RGB input. Each example shows the RGB image and its depth, alongside the Laplacian-transformed image and its corresponding depth.}
    \label{fig:privacy-preserving}\vspace{-1mm}
\end{figure}

\begin{figure}[t!] %\vspace{-5mm}
    \centering\setlength{\abovecaptionskip}{0.2cm}
    \includegraphics[width=0.48\textwidth]{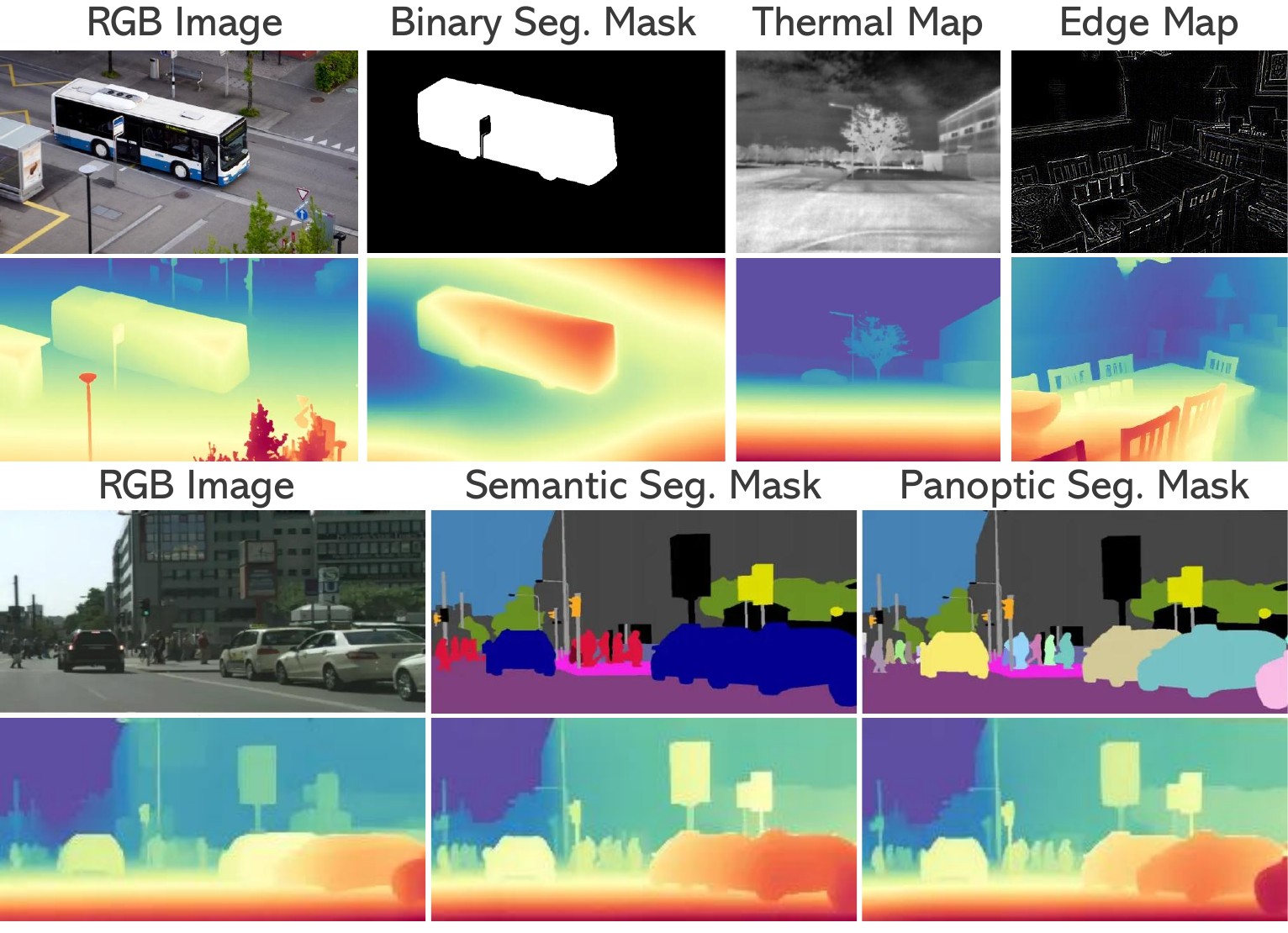}
\caption{\textbf{Generalization of depth prediction to diverse high-frequency visual prompts beyond Laplacian input}, including segmentation masks, thermal maps, and edge maps.}
    \label{fig:laplacian-invariance}\vspace{-1mm}
\end{figure}

\vspace{1mm}
\noindent\textbf{Privacy-preserving potential of Laplacian visual prompts in non-ambiguous scenes.} Fig.~\ref{fig:privacy-preserving} demonstrates that relative depth estimation using Laplacian-transformed visual prompts closely matches results obtained from standard RGB images in unambiguous scenes, highlighting LVP's strong generalization ability. This characteristic enables privacy-preserving applications by allowing only the Laplacian-transformed image to be transmitted to a cloud-based depth foundation model, thereby avoiding the transmission of sensitive low-level image features that could reveal human identities. Furthermore, Fig.~\ref{fig:laplacian-invariance} illustrates that this property extends to various high-frequency visual prompts. Combined with the spatial decoupling findings in our main paper, LVP ensures depth prediction invariance in unambiguous scenes while enabling spatial decoupling in ambiguous ones, advancing the development of ambiguity-aware spatial foundation models.

\section{More Quantitative Results}

 \begin{table}[t]
    \setlength{\abovecaptionskip}{0.2cm}
    \centering   \renewcommand{\arraystretch}{1.3}
    \resizebox{0.48\textwidth}{!}{
        \centering
        \setlength\tabcolsep{1.3mm}
         \begin{tabular}{l|c|c|cc}
            \hline \hline
             \textbf{Method} & \textbf{Semantics} & \textbf{\textit{Overall}} & \textbf{\textit{Reverse}} & \textbf{\textit{Same}} \\
           \hline 
          {LVP} (\textbf{Default}) & No & {\textbf{75.5}} & {\textbf{52.2}} & {\textbf{93.6}} \\ \hline 
                   \textcolor[HTML]{505050}{+ Predicted Mask ${}^{\dagger}$} & \textcolor[HTML]{505050}{Yes} & {\textcolor[HTML]{505050}{75.8}} & {\textcolor[HTML]{505050}{55.2}} & {\textcolor[HTML]{505050}{91.6}}
 \\
     \textcolor[HTML]{505050}{+ GT Mask (Ideal)} & {\textcolor[HTML]{505050}{Yes}} & {\textcolor[HTML]{505050}{82.5}} & {\textcolor[HTML]{505050}{69.2}} & {\textcolor[HTML]{505050}{92.7}} \\
            \hline  \hline
\multicolumn{5}{l}{\small $\dagger$ Overall predicted mask quality on \texttt{\texttt{MD-3k}}:  0.8815 in mean IoU.} \\
        \end{tabular}
    }
\caption{Even without using any extra semantic priors, our {LVP-empowered multi-layer depth} achieves comparable performance (measured by ML-SRA [\%]) to multi-layer depth interpolation using semantic priors (predicted mask from \cite{lin2021rich}).}
    \label{tab:comparison_LVP}\vspace{-2mm}
\end{table}

\vspace{0.5mm}
\noindent\textbf{LVP enables training-free multi-layer depth, approaching performance with semantic priors, highlighting its effectiveness.}
Table~\ref{tab:comparison_LVP} shows that the LVP-powered model (DAv2-L as baseline) achieves ML-SRA comparable to methods utilizing semantic priors, such as predicted segmentation masks. This demonstrates LVP's ability to leverage the model's implicit spatial understanding for spatial decoupling, without the need for domain-specific knowledge or post-processing, positioning it as a key component in developing generic MH-SFMs.

Our semantics-guided approach to multi-layer depth estimation integrates monocular depth predictions with semantic segmentation. We use the Depth Anything v1 Large model (DAv1-L)~\cite{depth_anything} for initial single-layer depth estimation. As noted in the main paper, DAv1-L tends to predict greater depths in ambiguous regions. Building on this bias, we estimate the nearer depth layer, typically corresponding to transparent surfaces, by interpolating depth values from the boundaries of ambiguous regions. This process is guided by a segmentation mask of the transparent surface and informed by DAv1-L's depth estimates outside the ambiguous regions.

Figures~\ref{fig:depth_seg_good} and~\ref{fig:depth_seg_bad} present qualitative results, showcasing both successful and failure cases. While this hybrid approach, combining DAv1-L’s depth bias with semantic segmentation, achieves higher quantitative precision for multi-layer depth estimation than our training-free LVP method, we emphasize the importance of developing foundational models that can directly handle multi-layer depth estimation, rather than relying on task-specific model combinations.

\begin{table*}[t]
    \centering
    \setlength{\abovecaptionskip}{0.2cm}
    \resizebox{\textwidth}{!}{
        \centering
        \renewcommand{\arraystretch}{1.30}
        \setlength\tabcolsep{1mm}
        \begin{tabular}{l|cccc|cccc|cc|cc}
            \hline \hline
            \multirow{3}{*}{\textbf{Baseline}} & \multicolumn{4}{c|}{\textbf{(a) \texttt{MD-3k} (\textit{Overall} Set)}} & \multicolumn{4}{c|}{\textbf{(b) \texttt{MD-3k} (\textit{Reverse} Subset)}} & \multicolumn{2}{c|}{\textbf{(c) \texttt{MD-3k} (\textit{Same} Subset)}} & \multicolumn{2}{c}{\textbf{(d)} \texttt{\textbf{DA-2k}}} \\ \cline{2-13}
             & \multicolumn{2}{c}{RGB Input} & \multicolumn{2}{c|}{LVP Input} & \multicolumn{2}{c}{RGB Input} & \multicolumn{2}{c|}{LVP Input} & RGB & LVP & RGB & LVP \\ \cline{2-13}
             &  \textcolor[HTML]{DB4437}{SRA(1)} & \textcolor[HTML]{4285F4}{SRA(2)} &  \textcolor[HTML]{DB4437}{SRA(1)} & \textcolor[HTML]{4285F4}{SRA(2)} &  \textcolor[HTML]{DB4437}{SRA(1)} & \textcolor[HTML]{4285F4}{SRA(2)} &  \textcolor[HTML]{DB4437}{SRA(1)} & \textcolor[HTML]{4285F4}{SRA(2)} & SRA(1/2) & SRA(1/2) & SRA & SRA \\
            \hline 
           Random & 50.0 & 50.0 & 50.0 & 50.0 & 50.0 & 50.0 & 50.0 & 50.0 & 50.0 & 50.0 & 50.0 & 50.0 \\ \hline 
            Marigold & 70.7 & \textcolor[HTML]{4285F4}{\textbf{79.9}} & \textcolor[HTML]{DB4437}{\textbf{70.8}}  & 77.4 & 39.6 & \textcolor[HTML]{4285F4}{\textbf{60.4}} & \textcolor[HTML]{DB4437}{\textbf{42.3}}  & 57.7 & 95.0 & 92.6 & 88.9 & 78.9 \\
         \rowcolor[HTML]{F1F3F4}   GeoWizard & 68.3 & \textcolor[HTML]{4285F4}{\textbf{83.8}} & \textcolor[HTML]{DB4437}{\textbf{70.3}}  & 78.7 & 32.3 & \textcolor[HTML]{4285F4}{\textbf{67.7}} & \textcolor[HTML]{DB4437}{\textbf{40.3}}  & 59.7 & 96.2 & 93.4 & 90.3 & 85.0 \\
            ZoeDepth & 58.7 & \textcolor[HTML]{4285F4}{\textbf{92.6}} & \textcolor[HTML]{DB4437}{\textbf{74.6}}  & 70.9 & 12.0 & \textcolor[HTML]{4285F4}{\textbf{88.0}} & \textcolor[HTML]{DB4437}{\textbf{54.3}}  & 45.7 & 94.9 & 90.4 & 86.7 & 77.2 \\
        \rowcolor[HTML]{F1F3F4}    DPT & 58.0 & \textcolor[HTML]{4285F4}{\textbf{91.9}} & \textcolor[HTML]{DB4437}{\textbf{75.7}}  & 71.1 & 10.2 & \textcolor[HTML]{4285F4}{\textbf{89.8}} & \textcolor[HTML]{DB4437}{\textbf{55.2}}  & 44.8 & 94.8 & 91.5 & 83.2 & 72.2 \\ \hline %\hline
            DAv1-S & 56.8 & \textcolor[HTML]{4285F4}{\textbf{95.3}} & \textcolor[HTML]{DB4437}{\textbf{60.6}}  & 85.4 & 5.8 & \textcolor[HTML]{4285F4}{\textbf{94.2}} & \textcolor[HTML]{DB4437}{\textbf{21.5}}  & 78.5 & 96.1 & 90.7 & 88.7 & 83.0 \\
       \rowcolor[HTML]{F1F3F4}     DAv1-B & 56.8 & \textcolor[HTML]{4285F4}{\textbf{95.4}} & \textcolor[HTML]{DB4437}{\textbf{58.8}}  & 89.7 & 5.8 & \textcolor[HTML]{4285F4}{\textbf{94.2}} & \textcolor[HTML]{DB4437}{\textbf{14.6}}  & 85.4 & 96.3 & 93.0 & 89.9 & 86.2 \\
            DAv1-L & 56.6 & \textcolor[HTML]{4285F4}{\textbf{96.0}} & \textcolor[HTML]{DB4437}{\textbf{59.2}}  & 90.9 & 4.8 & \textcolor[HTML]{4285F4}{\textbf{95.2}} & \textcolor[HTML]{DB4437}{\textbf{13.6}}  & 86.4 & 96.6 & 94.3 & 89.5 & 88.5 \\ \hline
            DAv2-O-S & 71.6 & \textcolor[HTML]{4285F4}{\textbf{77.3}} & \textcolor[HTML]{DB4437}{\textbf{81.4}}  & 60.1 & 43.4 & \textcolor[HTML]{4285F4}{\textbf{56.6}} & \textcolor[HTML]{DB4437}{\textbf{74.4}}  & 25.6 & 93.3 & 86.8 & 82.3 & 60.9 \\
        \rowcolor[HTML]{F1F3F4}    DAv2-O-B & 70.3 & \textcolor[HTML]{4285F4}{\textbf{80.3}} & \textcolor[HTML]{DB4437}{\textbf{77.2}}  & 65.8 & 38.5 & \textcolor[HTML]{4285F4}{\textbf{61.5}} & \textcolor[HTML]{DB4437}{\textbf{63.1}}  & 36.9 & 94.8 & 88.2 & 89.7 & 70.4 \\
            DAv2-O-L & 70.4 & \textcolor[HTML]{4285F4}{\textbf{82.4}} & \textcolor[HTML]{DB4437}{\textbf{71.5}}  & 79.5 & 36.2 & \textcolor[HTML]{4285F4}{\textbf{63.8}} & \textcolor[HTML]{DB4437}{\textbf{40.8}}  & 59.2 & 96.7 & 95.2 & 93.7 & 83.8 \\ \hline
            DAv2-I-S & \textcolor[HTML]{DB4437}{\textbf{80.4}} & 71.2 & {{70.0}} & \textcolor[HTML]{4285F4}{\textbf{74.3}}  & \textcolor[HTML]{DB4437}{\textbf{60.6}} & 39.4 & {{45.1}} & \textcolor[HTML]{4285F4}{\textbf{54.9}}  & 95.8 & 89.3 & 88.5 & 79.2 \\
       \rowcolor[HTML]{F1F3F4}     DAv2-I-B & \textcolor[HTML]{DB4437}{\textbf{83.1}} & 69.7 & {{72.6}} & \textcolor[HTML]{4285F4}{\textbf{76.1}}  & \textcolor[HTML]{DB4437}{\textbf{65.3}} & 34.7 & {{46.0}} & \textcolor[HTML]{4285F4}{\textbf{54.0}}  & 96.8 & 93.1 & 91.8 & 81.1 \\
            DAv2-I-L & \textcolor[HTML]{DB4437}{\textbf{85.2}} & 68.1 & {{68.1}} & \textcolor[HTML]{4285F4}{\textbf{82.6}}  & \textcolor[HTML]{DB4437}{\textbf{69.6}} & 30.4 & {{33.5}} & \textcolor[HTML]{4285F4}{\textbf{66.5}}  & 97.3 & 95.0 & 94.8 & 88.4 \\ \hline
            DAv2-S & \textcolor[HTML]{DB4437}{\textbf{78.0}} & 76.2 & {{61.5}} & \textcolor[HTML]{4285F4}{\textbf{85.8}}  & \textcolor[HTML]{DB4437}{\textbf{52.0}} & 48.0 & {{22.1}} & \textcolor[HTML]{4285F4}{\textbf{77.9}}  & 98.0 & 91.9 & 95.1 & 86.6 \\
      \rowcolor[HTML]{F1F3F4}      DAv2-B & \textcolor[HTML]{DB4437}{\textbf{82.4}} & 72.3 & {{60.5}} & \textcolor[HTML]{4285F4}{\textbf{88.6}}  & \textcolor[HTML]{DB4437}{\textbf{61.7}} & 38.3 & {{17.7}} & \textcolor[HTML]{4285F4}{\textbf{82.3}}  & 98.5 & 93.5 & 96.7 & 89.5 \\
            DAv2-L & \textcolor[HTML]{DB4437}{\textbf{84.0}} & 70.6 & {{60.2}} & \textcolor[HTML]{4285F4}{\textbf{89.9}}  & \textcolor[HTML]{DB4437}{\textbf{65.3}} & 34.7 & {{15.9}} & \textcolor[HTML]{4285F4}{\textbf{84.1}}  & 98.5 & 94.4 & 96.9 & 91.5 \\% \hline
            \hline \hline
        \end{tabular}
    }
    \caption{\textbf{Spatial Relationship Accuracy (SRA) [\%] for diverse baseline models on (a-c) the \texttt{MD-3k} benchmark and (d) the \texttt{DA-2k} dataset (non-ambiguous reference).} \textcolor[HTML]{DB4437}{SRA(1)} and \textcolor[HTML]{4285F4}{SRA(2)} measure the accuracy between the model's single-layer depth estimate and the  \textcolor[HTML]{DB4437}{first} and \textcolor[HTML]{4285F4}{second} layer annotation labels, respectively; higher values indicate a preference for that layer. Input modality (RGB or LVP) yielding higher \textcolor[HTML]{DB4437}{SRA(1)} or \textcolor[HTML]{4285F4}{SRA(2)}  per model on the full {\textit{overall}} set and {\textit{reverse}} subset of \texttt{MD-3k} are \textbf{bolded}.}
    \label{tab:detailed_benchmarking}% 
    \vspace{-2mm}
\end{table*}

\noindent\textbf{Detailed benchmarking results on \texttt{MD-3k}.}
Table~\ref{tab:detailed_benchmarking} presents per-layer Spatial Relationship Accuracy (SRA) [\%] for various baseline models on the \texttt{MD-3k} benchmark and the \texttt{DA-2k} dataset (non-ambiguous reference) for future comparison.

\begin{table}[t]
   \setlength{\abovecaptionskip}{0.2cm}
    \centering   \renewcommand{\arraystretch}{1.50}
    \resizebox{0.48\textwidth}{!}{
        \centering
        \setlength\tabcolsep{1mm}
         \begin{tabular}{l|c|ccc}% |>{\columncolor{Green!8}}c
            \hline \hline
             \textbf{Baseline} &  LVP (\textbf{Default}) & \textbf{LVP-2} & \textbf{LVP-R} & \textbf{LVP-G} \\
            \hline
           Random &25.0 & 25.0 & 25.0 &25.0 \\ \hline \hline
          %      \multicolumn{5}{c}{Relative  Depth Estimation} \\ \hline
            Marigold & {57.4}  &56.3 \textcolor[HTML]{DB4437}{(-1.1)}   &57.9 \textcolor[HTML]{4285F4}{(+0.5)}  &56.2 \textcolor[HTML]{DB4437}{(-1.2)} \\
     \rowcolor[HTML]{F1F3F4}       GeoWizard &59.5  &59.2 \textcolor[HTML]{DB4437}{(-0.3)}   &59.6 \textcolor[HTML]{4285F4}{(+0.1)}  &61.8 \textcolor[HTML]{4285F4}{(+2.3)} \\
            ZoeDepth & 68.8&67.0 \textcolor[HTML]{DB4437}{(-1.8)} &68.9 \textcolor[HTML]{4285F4}{(+0.1)} &65.1 \textcolor[HTML]{DB4437}{(-3.7)} \\
     \rowcolor[HTML]{F1F3F4}       DPT    & 70.2&68.5 \textcolor[HTML]{DB4437}{(-1.7)} &69.4 \textcolor[HTML]{DB4437}{(-0.8)} &68.6 \textcolor[HTML]{DB4437}{(-1.6)} \\ \hline \hline
            DAv1-S &57.9 &58.3 \textcolor[HTML]{4285F4}{(+0.4)} &57.7 \textcolor[HTML]{DB4437}{(-0.2)} &58.5 \textcolor[HTML]{4285F4}{(+0.6)} \\
     \rowcolor[HTML]{F1F3F4}       DAv1-B & 56.6 &56.9 \textcolor[HTML]{4285F4}{(+0.3)} &56.0 \textcolor[HTML]{DB4437}{(-0.6)} &56.3 \textcolor[HTML]{DB4437}{(-0.3)} \\
            DAv1-L & 57.1&56.9 \textcolor[HTML]{DB4437}{(-0.2)} &56.6 \textcolor[HTML]{DB4437}{(-0.5)} &57.4 \textcolor[HTML]{4285F4}{(+0.3)} \\ \hline
            DAv2-O-S & 63.0 &62.6 \textcolor[HTML]{DB4437}{(-0.4)} &62.8 \textcolor[HTML]{DB4437}{(-0.2)} &62.1 \textcolor[HTML]{DB4437}{(-0.9)} \\
     \rowcolor[HTML]{F1F3F4}       DAv2-O-B & 62.7&62.3 \textcolor[HTML]{DB4437}{(-0.4)} &62.6 \textcolor[HTML]{DB4437}{(-0.1)} &62.1 \textcolor[HTML]{DB4437}{(-0.6)} \\
            DAv2-O-L & 60.4&59.1 \textcolor[HTML]{DB4437}{(-1.3)} &60.4 \textcolor[HTML]{DB4437}{(-0.0)} &60.1 \textcolor[HTML]{DB4437}{(-0.3)} \\ \hline
            DAv2-I-S & 60.9 &61.9 \textcolor[HTML]{4285F4}{(+1.0)} &60.2 \textcolor[HTML]{DB4437}{(-0.7)} &58.6 \textcolor[HTML]{DB4437}{(-2.3)} \\
      \rowcolor[HTML]{F1F3F4}      DAv2-I-B & 63.7 &64.9 \textcolor[HTML]{4285F4}{(+1.2)} &63.7 \textcolor[HTML]{DB4437}{(-0.0)} &62.5 \textcolor[HTML]{DB4437}{(-1.2)} \\
            DAv2-I-L &{71.1} &{71.8} \textcolor[HTML]{4285F4}{(+0.7)} &{70.5} \textcolor[HTML]{DB4437}{(-0.6)} &{69.3} \textcolor[HTML]{DB4437}{(-1.8)} \\ \hline
            DAv2-S &67.2&68.1 \textcolor[HTML]{4285F4}{(+0.9)} &67.1 \textcolor[HTML]{DB4437}{(-0.1)} &67.2 \textcolor[HTML]{DB4437}{(-0.0)} \\
      \rowcolor[HTML]{F1F3F4}      DAv2-B &73.3&73.2 \textcolor[HTML]{DB4437}{(-0.1)} &73.0 \textcolor[HTML]{DB4437}{(-0.3)} &72.4 \textcolor[HTML]{DB4437}{(-0.9)} \\
            DAv2-L &\textbf{75.5} &\textbf{75.4} \textcolor[HTML]{DB4437}{(-0.1)} &\textbf{75.4} \textcolor[HTML]{DB4437}{(-0.1)} &\textbf{74.4} \textcolor[HTML]{DB4437}{(-1.1)} \\
            \hline  \hline
        \end{tabular}
    }%  
\caption{\textbf{Ablation study of LVP design} via ML-SRA [\%]. }
    \label{tab:ablation_comparison}\vspace{-4mm}
\end{table}

\noindent\textbf{Detailed ablation study results of Laplacian Visual Prompting design on \texttt{MD-3k}.}
Table~\ref{tab:ablation_comparison} presents ML-SRA performance under different Laplacian discretization (4-neighbor vs. 8-neighbor in default LVP and LVP-2), kernel sign (LVP-R with reversed convolution vs. LVP), and color space (RGB vs. grayscale, LVP vs. LVP-G). The results show that ML-SRA performance remains largely unaffected, highlighting the crucial role of high-frequency information in 3D spatial decoupling.

\section{Implementation Details}

\subsection{Laplacian Visual Prompting for Sequential Video Inference}

We introduce Laplacian Visual Prompting (LVP) for sequential video inference, specifically for multi-layer depth estimation. Using the Video Depth Anything model as our baseline, LVP follows the multi-hypothesis depth estimation framework outlined in the main paper. Crucially, LVP enhances temporal consistency in depth estimation, mitigating inconsistencies common in per-frame processing. Additionally, the extracted hidden depth maps maintain high fidelity throughout the video sequence. A \textbf{video demonstration} is provided in the supplementary material.

\subsection{Depth-Conditioned Image Generation}

Our image generation pipeline uses depth maps and text prompts to guide scene synthesis. Depth maps are estimated using our proposed method, ensuring geometric accuracy, while text prompts modulate the visual attributes of the generated images.

We employ two distinct text prompts to control the scene's appearance:

\begin{itemize}
    \item \texttt{a bright, well-lit photograph of an interior space with natural daylight, clear windows, balanced lighting, accurate geometry and structure, photorealistic, vibrant colors, modern interior design, clean and airy space}
\end{itemize}

\begin{figure*}[t!]
\centering \setlength{\abovecaptionskip}{0.1cm}
\includegraphics[width=\textwidth]{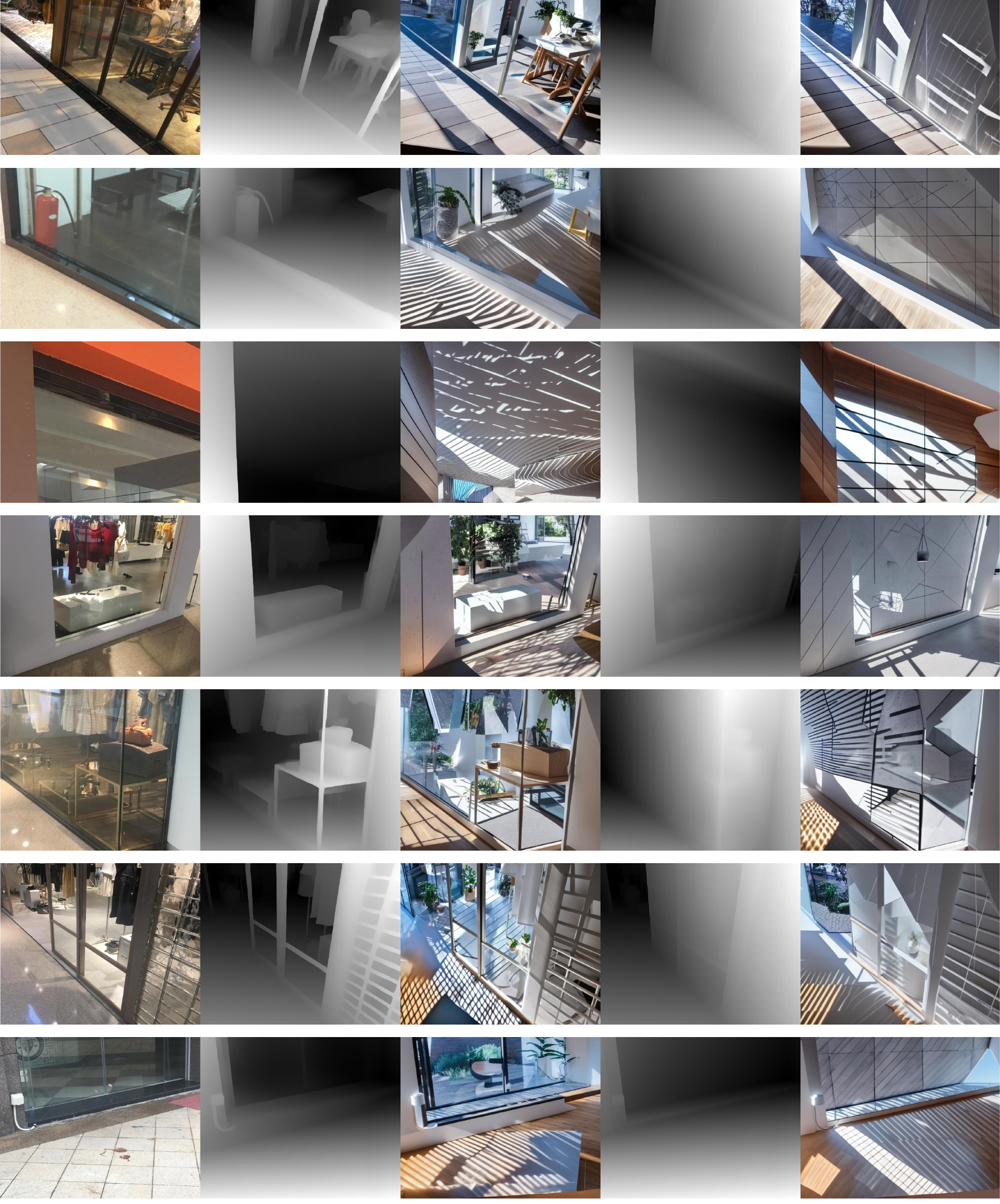}
\caption{\textbf{Multi-hypothesis spatial understanding enhances flexible geometry-conditioned visual generation.} From left to right: original RGB image, depth from Laplacian Visual Prompting with its corresponding generated RGB image, and depth from the original RGB image with its generated RGB counterpart. }
\label{fig:visual_gen_appendix_1}%\vspace{-2mm}
\end{figure*}

The resulting RGB images, synthesized from multi-layer depth representations derived via LVP, are illustrated in Fig.~\ref{fig:visual_gen_appendix_1}.

\begin{itemize}
    \item \texttt{a bright, well-lit photograph of an interior space, accurate geometry and structure, photorealistic, modern interior design, clean and airy space}
\end{itemize}

\begin{figure*}[t!]
\centering \setlength{\abovecaptionskip}{0.1cm}
\includegraphics[width=\textwidth]{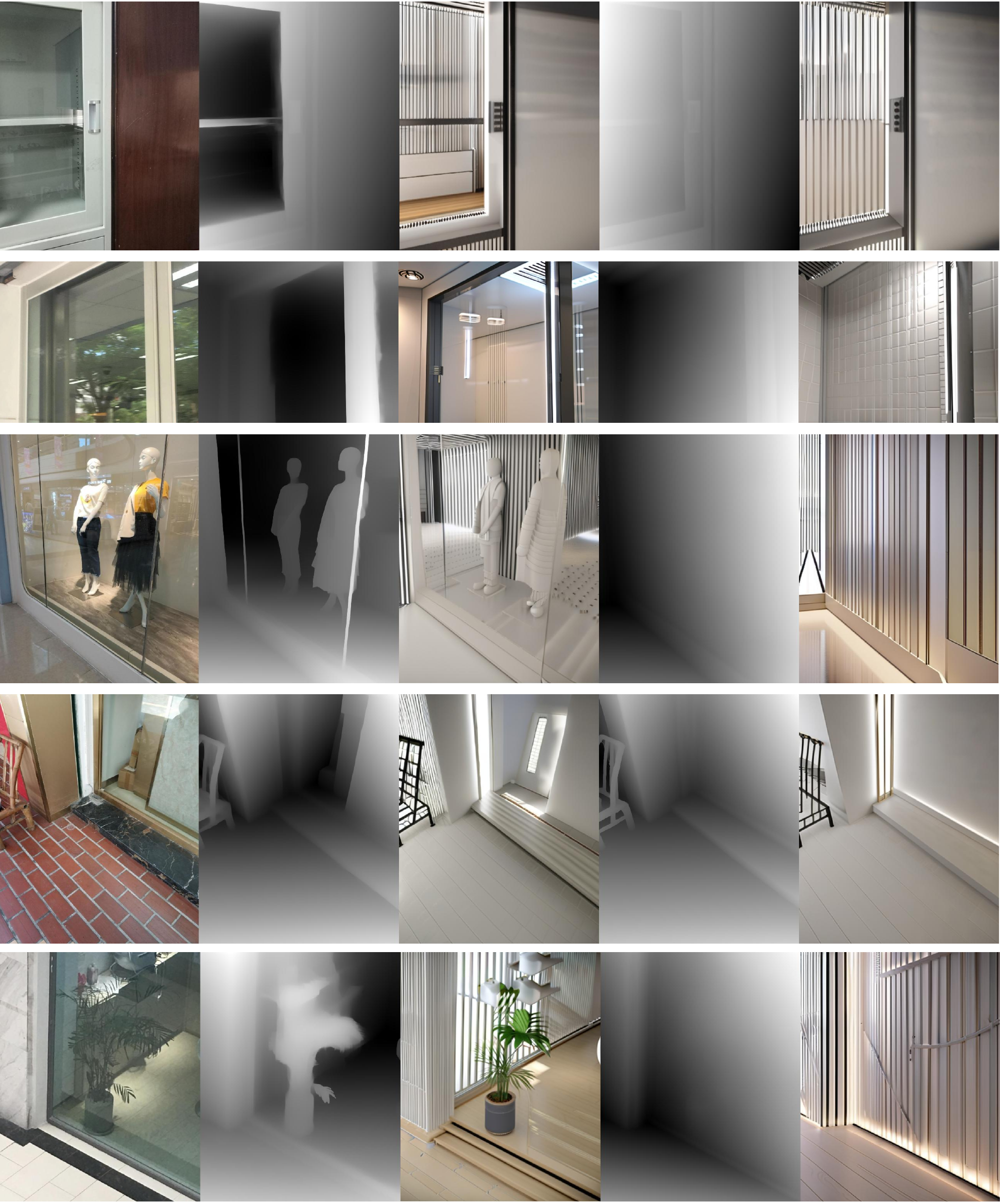}
\caption{\textbf{Multi-hypothesis spatial understanding enhances flexible geometry-conditioned visual generation.} From left to right: original RGB image, depth from Laplacian Visual Prompting with its corresponding generated RGB image, and depth from the original RGB image with its generated RGB counterpart. }
\label{fig:visual_gen_appendix_2}%\vspace{-2mm}
\end{figure*}

The corresponding RGB images, generated using multi-layer depth representations from LVP, are presented in Fig.~\ref{fig:visual_gen_appendix_2}.

These prompts enhance photorealistic scene synthesis by ensuring precise geometry and well-balanced lighting, complementing depth information. More details on the diffusion models used for depth-conditioned visual generation are available in the \textit{Diffusers} library~\footnote{\url{https://huggingface.co/docs/diffusers/en/index}}.

\section{\textcolor[rgb]{0.0,0.0,0.0}{Datasheet for \textit{\texttt{MD-3k}} Benchmark}}
\label{sec:datasheet}

We document the necessary information about the proposed datasets and benchmarks following the guidelines of Gebru \textit{et al}.~\citep{datasheet}.

\begin{enumerate}[label=Q\arabic*]

\vspace{-0.5em}\begin{figure}[!h]
\subsection{\textcolor[rgb]{0.0,0.0,0.0}{Motivation}}\label{subsec:datasheet-motivation}\vspace{-1em}
\end{figure}

\begin{figure}[t!]
    \centering
    \setlength{\abovecaptionskip}{0.1cm}
    \includegraphics[width=0.48\textwidth]{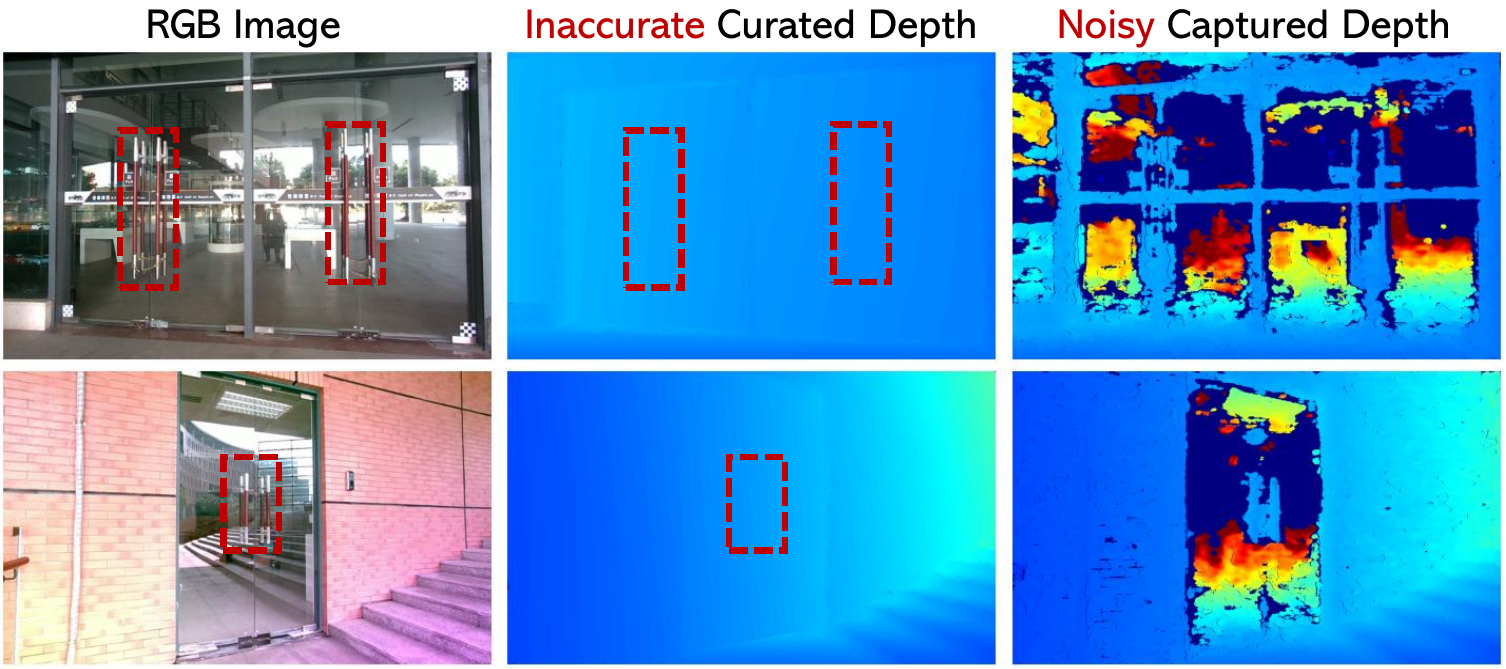}
\caption{Noise and inaccuracies in depth data from existing datasets~\cite{liang2023monocular} for complex, ambiguous scenes. These arise from limitations in both sensor acquisition and human annotation.}
    \label{fig:challenge_existing_dataset}
\end{figure}

\item \textbf{For what purpose was the dataset created?} Was there a specific task in mind? Was there a specific gap that needed to be filled? Please provide a description.

\begin{itemize}
\item Fig.~\ref{fig:challenge_existing_dataset} highlights the limitations of existing datasets for ambiguous transparent scenes.  They often contain noisy raw depth from sensors (due to physical limitations) and inaccurate curated depth (due to human error). These challenges motivate our creation of the \texttt{MD-3k} benchmark.

\item Our benchmark was created to evaluate multi-layer spatial perception, specifically focusing on the challenge of depth disentanglement in ambiguous 3D scenes. Existing depth datasets lack multi-layer spatial relationship labels, hindering fine-grained analysis in regions with transparency and spatial ambiguity. \texttt{MD-3k} fills this gap by providing the first benchmark with multi-layer spatial relationship labels, enabling detailed evaluation of models' ability to understand layered depth ordering.
\end{itemize}

\item \textbf{Who created the dataset (e.g., which team, research group) and on behalf of which entity (e.g., company, institution, organization)?}

\begin{itemize}
\item This benchmark is presented by [Anonymous Author]\footnote{Author and repository information have been anonymized in compliance with the double-blind policy and will be provided upon acceptance.}. Our aim is to advance the development and evaluation of spatial perception models, particularly in understanding multi-layer depth relationships in complex scenes.
\end{itemize}

\item \textbf{Who funded the creation of the dataset?} If there is an associated grant, please provide the name of the grantor and the grant name and number.

\begin{itemize}
\item This work was partially supported by [Anonymous Author].
\end{itemize}

\item \textbf{Any other comments?}

\begin{itemize}
\item No.
\end{itemize}

\vspace{-0.5em}\begin{figure}[!h]
\subsection{\textcolor[rgb]{0.0,0.0,0.0}{Composition}}\label{subsec:datasheet-composition}\vspace{-1em}
\end{figure}

\item \textbf{What do the instances that comprise the dataset represent (e.g., documents, photos, people, countries)?} \textit{Are there multiple types of instances (e.g., movies, users, and ratings; people and interactions between them; nodes and edges)? Please provide a description.}

\begin{itemize}
\item The instances in \texttt{MD-3k} represent high-resolution RGB images of indoor and outdoor scenes containing ambiguous regions, particularly those involving transparent objects. Each instance is associated with segmentation masks highlighting ambiguous regions and pairwise spatial relationship labels for sparse points within these regions.
\end{itemize}

\item \textbf{How many instances are there in total (of each type, if appropriate)?}

\begin{itemize}
\item The \texttt{MD-3k} benchmark comprises 3,161 high-resolution RGB images. Each image contains annotations of spatial relationships for pairs of sparse points in ambiguous regions, totaling 3,161 annotated pairs. For each pair, two spatial relationship labels are provided, representing the relationship `on' and another for the region `behind' the transparent object.
\end{itemize}

\item \textbf{Does the dataset contain all possible instances or is it a sample (not necessarily random) of instances from a larger set?} \textit{If the dataset is a sample, what is the larger set? Is the sample representative of the larger set (e.g., geographic coverage)? If so, please describe how this representativeness was validated/verified. If it is not representative of the larger set, please describe why not (e.g., to cover a more diverse range of instances, because instances were withheld or unavailable).}

\begin{itemize}
\item The dataset is a carefully selected sample from the GDD segmentation dataset~\citep{mei2020don}. The larger set is the entire GDD dataset.  The sample is not random but specifically chosen to include scenes rich in ambiguous regions, particularly those with transparent objects, to address the benchmark's focus on multi-layer spatial understanding in such challenging scenarios.
\end{itemize}

\item \textbf{What data does each instance consist of?} \textit{“Raw” data (e.g., unprocessed text or images) or features? In either case, please provide a description.}

\begin{itemize}
\item Each instance consists of:
\begin{itemize}
    \item \textbf{RGB image:} High-resolution (720p) RGB image in PNG format.
    \item \textbf{Segmentation masks:} Binary masks highlighting ambiguous regions within the RGB image, in PNG format.
    \item \textbf{Spatial relationship labels:} Pairwise spatial relationship labels for sparse points in ambiguous regions, provided in JSON format. Each pair has two labels: one for the region `on' and another for the region `behind' the transparent object, indicating depth ordering (above and behind).
\end{itemize}
This data is considered `raw' in the sense that it is primarily image data and annotations, not pre-extracted features.
\end{itemize}

\item \textbf{Is there a label or target associated with each instance?} \textit{If so, please provide a description.}

\begin{itemize}
\item Yes, the primary labels are the \textbf{pairwise spatial relationship labels}. For each annotated pair of sparse points in an ambiguous region of an RGB image, there are two labels indicating the spatial relationship (depth order) between the points in two layers: one for the region `on' and another for the region `behind' the transparent object. The labels are `above' and `behind'.
\end{itemize}

\item \textbf{Is any information missing from individual instances?} \textit{If so, please provide a description, explaining why this information is missing (e.g., because it was unavailable). This does not include intentionally removed information, but might include, e.g., redacted text.}

\begin{itemize}
\item No, all instances are fully annotated with RGB images, segmentation masks, and spatial relationship labels.
\end{itemize}

\item \textbf{Are relationships between individual instances made explicit (e.g., users' movie ratings, social network links)?} \textit{If so, please describe how these relationships are made explicit.}

\begin{itemize}
\item The instances are related by their source dataset, GDD~\citep{mei2020don}. All images are selected from GDD and share the characteristics of scenes within that dataset.  Furthermore, images are implicitly related by the common theme of containing ambiguous regions and transparent objects, as this was the selection criterion.
\end{itemize}

\item \textbf{Are there recommended data splits (e.g., training, development/validation, testing)?} \textit{If so, please provide a description of these splits, explaining the rationale behind them.}

\begin{itemize}
\item No, we do not provide predefined data splits. Users are free to define their own splits based on their specific research needs. Common splits like training/validation/testing can be created randomly or based on scene characteristics if desired.
\end{itemize}

\item \textbf{Are there any errors, sources of noise, or redundancies in the dataset?} \textit{If so, please provide a description.}

\begin{itemize}
\item We have implemented a rigorous annotation pipeline, including multi-round verification by expert annotators, to minimize errors and noise in the spatial relationship labels. However, as with any human annotation, there might be minor inconsistencies or subjective interpretations. We believe the overall quality of the annotations is high due to the careful curation process. Redundancies are not intentionally introduced.
\end{itemize}

\item \textbf{Is the dataset self-contained, or does it link to or otherwise rely on external resources (e.g., websites, tweets, other datasets)?} \textit{If it links to or relies on external resources, a) are there guarantees that they will exist, and remain constant, over time; b) are there official archival versions of the complete dataset (i.e., including the external resources as they existed at the time the dataset was created); c) are there any restrictions (e.g., licenses, fees) associated with any of the external resources that might apply to a future user? Please provide descriptions of all external resources and any restrictions associated with them, as well as links or other access points, as appropriate.}

\begin{itemize}
\item The \texttt{MD-3k} benchmark is distributed as a self-contained dataset of annotations, segmentation masks, and image lists. It relies on the images from the GDD dataset~\citep{mei2020don} as the underlying visual data. Users will need to obtain the GDD dataset separately to use \texttt{MD-3k} fully.
    \begin{itemize}
        \item a) We cannot guarantee the long-term availability of the GDD dataset. However, GDD is a publicly available dataset for research purposes.
        \item b) We do not provide archival versions of the GDD dataset. Users should refer to the original GDD dataset sources for archival information.
        \item c) Users should adhere to the licensing terms of the GDD dataset, which are separate from the \texttt{MD-3k} benchmark license. Please refer to the GDD dataset documentation for details on licenses and restrictions.
    \end{itemize}
\end{itemize}

\item \textbf{Does the dataset contain data that might be considered confidential (e.g., data that is protected by legal privilege or by doctor–patient confidentiality, data that includes the content of individuals’ non-public communications)?} \textit{If so, please provide a description.}

\begin{itemize}
\item No, the \texttt{MD-3k} benchmark utilizes images from the publicly available GDD dataset, which does not contain confidential information.
\end{itemize}

\item \textbf{Does the dataset contain data that, if viewed directly, might be offensive, insulting, threatening, or might otherwise cause anxiety?} \textit{If so, please describe why.}

\begin{itemize}
\item No, the images in the \texttt{MD-3k} benchmark depict common indoor and outdoor scenes and do not contain offensive, insulting, threatening, or anxiety-inducing content to the best of our knowledge.
\end{itemize}

\item \textbf{Does the dataset relate to people?} \textit{If not, you may skip the remaining questions in this section.}

\begin{itemize}
\item No. This dataset primarily focuses on scenes and spatial relationships, and does not directly relate to people in a way that raises privacy or ethical concerns. While people may be present in some images as part of the general scene context, the annotations and benchmark tasks are not centered around individuals.
\end{itemize}

\item \textbf{Does the dataset identify any subpopulations (e.g., by age, gender)?}

\begin{itemize}
\item N/A.
\end{itemize}

\item \textbf{Is it possible to identify individuals (i.e., one or more natural persons), either directly or indirectly (i.e., in combination with other data) from the dataset?} \textit{If so, please describe how.}

\begin{itemize}
\item N/A.
\end{itemize}

\item \textbf{Does the dataset contain data that might be considered sensitive in any way (e.g., data that reveals racial or ethnic origins, sexual orientations, religious beliefs, political opinions or union memberships, or locations; financial or health data; biometric or genetic data; forms of government identification, such as social security numbers; criminal history)?} \textit{If so, please provide a description.}

\begin{itemize}
\item No.
\end{itemize}

\item \textbf{Any other comments?}

\begin{itemize}
\item We encourage users to exercise discretion and use the benchmark responsibly for research purposes only.
\end{itemize}

\vspace{-0.5em}\begin{figure}[!h]
\subsection{\textcolor[rgb]{0.0,0.0,0.0}{Collection Process}}\label{subsec:datasheet-collection}\vspace{-1em}
\end{figure}

\item \textbf{How was the data associated with each instance acquired?} \textit{Was the data directly observable (e.g., raw text, movie ratings), reported by subjects (e.g., survey responses), or indirectly inferred/derived from other data (e.g., part-of-speech tags, model-based guesses for age or language)? If data was reported by subjects or indirectly inferred/derived from other data, was the data validated/verified? If so, please describe how.}

\begin{itemize}
\item The RGB images were directly observable, sourced from the GDD segmentation dataset~\citep{mei2020don}. The segmentation masks and spatial relationship labels were indirectly derived through expert human annotation. Expert annotators manually identified ambiguous regions and provided pairwise spatial relationship labels. The annotations were validated through a multi-round verification process involving multiple annotators to ensure consistency and accuracy.
\end{itemize}

\item \textbf{What mechanisms or procedures were used to collect the data (e.g., hardware apparatus or sensor, manual human curation, software program, software API)?} \textit{How were these mechanisms or procedures validated?}

\begin{itemize}
\item The data collection process primarily involved \textbf{manual human curation}. Expert annotators used in-house annotation tools to:
    \begin{itemize}
        \item Visually inspect RGB images from the GDD dataset.
        \item Identify and segment ambiguous regions, particularly those involving transparent objects, creating segmentation masks.
        \item Select sparse point pairs within these ambiguous regions.
        \item Determine and assign pairwise spatial relationship labels (`above' and `behind') for each pair, considering both layers (`on' and `behind' the transparent object).
    \end{itemize}
The annotation procedure was validated through a multi-round verification process. Different annotators reviewed and cross-validated annotations to resolve discrepancies and ensure consistency and accuracy of the labels.
\end{itemize}

\item \textbf{If the dataset is a sample from a larger set, what was the sampling strategy (e.g., deterministic, probabilistic with specific sampling probabilities)?}

\begin{itemize}
\item The sampling strategy was \textbf{deterministic and targeted}. Images were selected from the GDD dataset based on a specific criterion: the presence of ambiguous regions, especially those featuring transparent objects. This was a deliberate selection process to focus the benchmark on the challenges of multi-layer spatial understanding in such complex scenes, rather than a random or probabilistic sampling of the entire GDD dataset.
\end{itemize}

\item \textbf{Who was involved in the data collection process (e.g., students, crowdworkers, contractors) and how were they compensated (e.g., how much were crowdworkers paid)?}

\begin{itemize}
\item The data collection process involved \textbf{expert annotators}. These were individuals with expertise in computer vision and image annotation, specifically trained for the task of identifying ambiguous regions and determining spatial relationships. Compensation details are proprietary to [Anonymous Author] and are not disclosed for anonymity purposes.
\end{itemize}

\item \textbf{Over what timeframe was the data collected? Does this timeframe match the creation timeframe of the data associated with the instances (e.g., recent crawl of old news articles)?} \textit{If not, please describe the timeframe in which the data associated with the instances was created.}

\begin{itemize}
\item The data annotation and collection process took place between [Anonymous Month, Year] and [Anonymous Month, Year]. This timeframe represents the creation timeframe of the segmentation masks and spatial relationship labels associated with the images from the GDD dataset. The GDD dataset itself was created prior to this timeframe.
\end{itemize}

\item \textbf{Were any ethical review processes conducted (e.g., by an institutional review board)?} \textit{If so, please provide a description of these review processes, including the outcomes, as well as a link or other access point to any supporting documentation.}

\begin{itemize}
\item Ethical review processes were not formally conducted by an institutional review board specifically for the creation of \texttt{MD-3k}. However, the benchmark utilizes publicly available images from the GDD dataset, which is intended for research purposes. We have taken care to ensure that the annotation process and the resulting benchmark do not raise ethical concerns regarding privacy or sensitive information.
\end{itemize}

\item \textbf{Does the dataset relate to people?} \textit{If not, you may skip the remaining questions in this section.}

\begin{itemize}
\item No.
\end{itemize}

\item \textbf{Did you collect the data from the individuals in question directly, or obtain it via third parties or other sources (e.g., websites)?}

\begin{itemize}
\item N/A.
\end{itemize}

\item \textbf{Were the individuals in question notified about the data collection?} \textit{If so, please describe (or show with screenshots or other information) how notice was provided, and provide a link or other access point to, or otherwise reproduce, the exact language of the notification itself.}

\begin{itemize}
\item N/A.
\end{itemize}

\item \textbf{Did the individuals in question consent to the collection and use of their data?} \textit{If so, please describe (or show with screenshots or other information) how consent was requested and provided, and provide a link or other access point to, or otherwise reproduce, the exact language to which the individuals consented.}

\begin{itemize}
\item N/A.
\end{itemize}

\item \textbf{If consent was obtained, were the consenting individuals provided with a mechanism to revoke their consent in the future or for certain uses?} \textit{If so, please provide a description, as well as a link or other access point to the mechanism (if appropriate).}

\begin{itemize}
\item N/A.
\end{itemize}

\item \textbf{Has an analysis of the potential impact of the dataset and its use on data subjects (e.g., a data protection impact analysis) been conducted?} \textit{If so, please provide a description of this analysis, including the outcomes, as well as a link or other access point to any supporting documentation.}

\begin{itemize}
\item No formal data protection impact analysis has been conducted. However, as the dataset does not directly relate to people and uses publicly available images, we anticipate minimal risk to data subjects. The focus of the benchmark is on evaluating computer vision models for spatial understanding. We encourage responsible use of the benchmark for research purposes.
\end{itemize}

\item \textbf{Any other comments?}

\begin{itemize}
\item No.
\end{itemize}

\vspace{-0.5em}\begin{figure}[!h]
\subsection{\textcolor[rgb]{0.0,0.0,0.0}{Preprocessing, Cleaning, and/or Labeling}}\label{subsec:datasheet-preprocess}
\vspace{-1em}
\end{figure}

\item \textbf{Was any preprocessing/cleaning/labeling of the data done (e.g., discretization or bucketing, tokenization, part-of-speech tagging, SIFT feature extraction, removal of instances, processing of missing values)?} \textit{If so, please provide a description. If not, you may skip the remainder of the questions in this section.}

\begin{itemize}
\item Yes, \textbf{labeling} was performed. Expert annotators manually labeled spatial relationships for pairs of sparse points in ambiguous regions. This labeling process is the core contribution of the \texttt{MD-3k} benchmark. No other preprocessing or cleaning of the RGB images from the GDD dataset was performed.
\end{itemize}

\item \textbf{Was the “raw” data saved in addition to the preprocessed/cleaned/labeled data (e.g., to support unanticipated future uses)?} \textit{If so, please provide a link or other access point to the “raw” data.}

\begin{itemize}
\item N/A. The `raw' data in this context would be the original RGB images from the GDD dataset. We are distributing the segmentation masks and spatial relationship labels, which are the `labeled' data. The `raw' RGB images are available from the original GDD dataset~\citep{mei2020don}.
\end{itemize}

\item \textbf{Is the software used to preprocess/clean/label the instances available?} \textit{If so, please provide a link or other access point.}

\begin{itemize}
\item The in-house annotation tools used for labeling are not publicly released at this time due to proprietary reasons. However, we will provide detailed descriptions of the annotation process and data format to facilitate the use of the benchmark.
\end{itemize}

\item \textbf{Any other comments?}

\begin{itemize}
\item No.
\end{itemize}

\vspace{-0.5em}\begin{figure}[!h]
\subsection{\textcolor[rgb]{0.0,0.0,0.0}{Uses}}\label{subsec:datasheet-uses}
\vspace{-1em}
\end{figure}

\item \textbf{Has the dataset been used for any tasks already?} \textit{If so, please provide a description.}

\begin{itemize}
\item Not yet. \texttt{MD-3k} is a newly introduced benchmark. We will present initial baseline evaluations in our paper.
\end{itemize}

\item \textbf{Is there a repository that links to any or all papers or systems that use the dataset?} \textit{If so, please provide a link or other access point.}

\begin{itemize}
\item We will maintain a repository at [Anonymous Repo] that links to papers and systems that utilize the \texttt{MD-3k} benchmark as they become available.
\end{itemize}

\item \textbf{What (other) tasks could the dataset be used for?}

\begin{itemize}
\item The primary intended use of \texttt{MD-3k} is for evaluating models for \textbf{multi-layer spatial understanding} and \textbf{depth disentanglement} in ambiguous scenes. Specifically, it can be used to:
    \begin{itemize}
        \item Evaluate the performance of depth estimation models in regions with transparency and complex spatial arrangements.
        \item Benchmark algorithms designed for understanding layered scene representations.
        \item Analyze the ability of models to reason about relative depth ordering in multi-layer contexts.
        \item Develop and test novel approaches for handling spatial ambiguity in 3D scene understanding.
    \end{itemize}
It can also be used for related tasks such as transparent object segmentation and reasoning about spatial relationships in general.
\end{itemize}

\item \textbf{Is there anything about the composition of the dataset or the way it was collected and preprocessed/cleaned/labeled that might impact future uses?} \textit{For example, is there anything that a future user might need to know to avoid uses that could result in unfair treatment of individuals or groups (e.g., stereotyping, quality of service issues) or other undesirable harms (e.g., financial harms, legal risks)? If so, please provide a description. Is there anything a future user could do to mitigate these undesirable harms?}

\begin{itemize}
\item The \texttt{MD-3k} benchmark is focused on ambiguous scenes, particularly those with transparent objects. Users should be aware that the dataset is specifically designed to challenge models in these scenarios.  It might not be representative of general scenes without ambiguity.  Future users should consider this focus when applying the benchmark and interpreting results.  As the dataset does not relate to people or sensitive attributes, the risk of unfair treatment or other harms is considered low. However, responsible and ethical use of the benchmark is always encouraged.
\end{itemize}

\item \textbf{Are there tasks for which the dataset should not be used?} \textit{If so, please provide a description.}

\begin{itemize}
\item We are not aware of any specific tasks for which \texttt{MD-3k} should not be used. However, its primary focus is on multi-layer spatial understanding in ambiguous regions. Using it for tasks completely unrelated to spatial reasoning or depth perception might not be appropriate.
\end{itemize}

\item \textbf{Any other comments?}

\begin{itemize}
\item No.
\end{itemize}

\vspace{-0.5em}\begin{figure}[!h]
\subsection{\textcolor[rgb]{0.0,0.0,0.0}{Distribution and License}}\label{subsec:datasheet-distribution}
\vspace{-1em}
\end{figure}

\item \textbf{Will the dataset be distributed to third parties outside of the entity (e.g.,
 company, institution, organization) on behalf of which the dataset was created?} \textit{If so, please provide a description.}

\begin{itemize}
\item Yes, the \texttt{MD-3k} benchmark will be publicly distributed to third parties for research purposes.
\end{itemize}

\item \textbf{How will the dataset be distributed (e.g., tarball on website, API, GitHub)?} \textit{Does the dataset have a digital object identifier (DOI)?}

\begin{itemize}
\item The \texttt{MD-3k} benchmark, including annotations, segmentation masks, and code for evaluation, will be distributed as a downloadable tarball via [Anonymous Repo], likely a GitHub repository. We plan to obtain a DOI for the benchmark upon publication of the associated paper.
\end{itemize}

\item \textbf{When will the dataset be distributed?}

\begin{itemize}
\item We plan to distribute the dataset publicly starting from March, 2025, and onwards, coinciding with the anticipated publication of our work.
\end{itemize}

\item \textbf{Will the dataset be distributed under a copyright or other intellectual property (IP) license, and/or under applicable terms of use (ToU)?} \textit{If so, please describe this license and/or ToU, and provide a link or other access point to, or otherwise reproduce, any relevant licensing terms or ToU, as well as any fees associated with these restrictions.}

\begin{itemize}
\item The \texttt{MD-3k} benchmark, including annotations and code, will be released under the \textbf{Apache-2.0 license}. This is an open-source license that allows for free use, modification, and distribution for research and commercial purposes, with proper attribution. The full license text will be available in the benchmark repository at [Anonymous Repo]. There are no fees associated with the use of the \texttt{MD-3k} benchmark under this license.
\end{itemize}

\item \textbf{Have any third parties imposed IP-based or other restrictions on the data associated with the instances?} \textit{If so, please describe these restrictions, and provide a link or other access point to, or otherwise reproduce, any relevant licensing terms, as well as any fees associated with these restrictions.}

\begin{itemize}
\item The \texttt{MD-3k} benchmark relies on RGB images from the GDD dataset~\citep{mei2020don}. Users of \texttt{MD-3k} should also comply with the licensing terms of the GDD dataset, which are separate from the Apache-2.0 license of our benchmark. We recommend users refer to the GDD dataset documentation for details on their specific licensing terms and any potential restrictions. We are not aware of any IP-based or other restrictions imposed by third parties directly on our annotations and benchmark data, other than the underlying GDD images.
\end{itemize}

\item \textbf{Do any export controls or other regulatory restrictions apply to the dataset or to individual instances?} \textit{If so, please describe these restrictions, and provide a link or other access point to, or otherwise reproduce, any supporting documentation.}

\begin{itemize}
\item No, we are not aware of any export controls or other regulatory restrictions applicable to the \texttt{MD-3k} benchmark or its individual instances.
\end{itemize}

\item \textbf{Any other comments?}

\begin{itemize}
\item No.
\end{itemize}

\vspace{-0.5em}\begin{figure}[!h]
\subsection{\textcolor[rgb]{0.0,0.0,0.0}{Maintenance}}\label{subsec:datasheet-mainteanance}
\vspace{-1em}
\end{figure}

\item \textbf{Who will be supporting/hosting/maintaining the dataset?}

\begin{itemize}
\item Anonymous Author will be responsible for supporting, hosting, and maintaining the \texttt{MD-3k} benchmark.
\end{itemize}

\item \textbf{How can the owner/curator/manager of the dataset be contacted (e.g., email address)?}

\begin{itemize}
\item The owner/curator/manager of the dataset can be contacted through [Anonymous Contact Method], which will be provided in the benchmark repository at [Anonymous Repo] upon release.
\end{itemize}

\item \textbf{Is there an erratum?} \textit{If so, please provide a link or other access point.}

\begin{itemize}
\item There is no erratum for the initial release of \texttt{MD-3k}. Any errata identified in the future will be documented and communicated through the benchmark repository at [Anonymous Repo].
\end{itemize}

\item \textbf{Will the dataset be updated (e.g., to correct labeling errors, add new instances, delete instances)?} \textit{If so, please describe how often, by whom, and how updates will be communicated to users (e.g., mailing list, GitHub)?}

\begin{itemize}
\item Yes, we plan to update the \texttt{MD-3k} benchmark periodically. Updates may include corrections of labeling errors, addition of new instances, or improvements to the benchmark resources. Updates will be performed by [Anonymous Author] and will be communicated to users through the benchmark repository at [Anonymous Repo], potentially via release notes and/or announcements on the repository's issue tracker.
\end{itemize}

\item \textbf{If the dataset relates to people, are there applicable limits on the retention of the data associated with the instances (e.g., were individuals in question told that their data would be retained for a fixed period of time and then deleted)?} \textit{If so, please describe these limits and explain how they will be enforced.}

\begin{itemize}
\item N/A.
\end{itemize}

\item \textbf{Will older versions of the dataset continue to be supported/hosted/maintained?} \textit{If so, please describe how. If not, please describe how its obsolescence will be communicated to users.}

\begin{itemize}
\item We intend to host and maintain older versions of the \texttt{MD-3k} benchmark in the benchmark repository at [Anonymous Repo], likely through version control mechanisms (e.g., Git tags or branches). This will allow users to access and utilize specific versions of the benchmark for reproducibility and comparison purposes.  If a version becomes obsolete, it will be clearly marked as such in the repository, but will remain accessible.
\end{itemize}

\item \textbf{If others want to extend/augment/build on/contribute to the dataset, is there a mechanism for them to do so?} \textit{If so, please provide a description. Will these contributions be validated/verified? If so, please describe how. If not, why not? Is there a process for communicating/distributing these contributions to other users? If so, please provide a description.}

\begin{itemize}
\item We welcome contributions from the community to extend, augment, or build upon the \texttt{MD-3k} benchmark. Users can contribute by:
    \begin{itemize}
        \item Reporting issues or suggesting improvements via the issue tracker in the benchmark repository at [Anonymous Repo].
        \item Submitting pull requests with code contributions (e.g., evaluation scripts, new baselines).
        \item Proposing new annotations or extensions to the dataset by contacting [Anonymous Author] through the contact method provided in the repository.
    \end{itemize}
\end{itemize}

\item \textbf{Any other comments?}

\begin{itemize}
\item No.
\end{itemize}

\end{enumerate}

\section{Broader Impact}\label{sec:broader_impact}

This work transcends monocular depth estimation, tackling the fundamental challenge of depth ambiguity that limits spatial understanding across AI. By resolving this core limitation, we pave the way for more reliable and versatile AI in complex 3D environments, impacting diverse fields reliant on robust perception.

Our central contribution, Laplacian Visual Prompting (LVP), introduces a training-free spectral technique. LVP empowers models to explicitly disentangle depth ambiguity and generate multi-hypothesis predictions, unlocking latent spatial knowledge. This broadly applicable spectral prompting paradigm extends beyond depth, offering a transformative tool for visual model adaptation and interpretation across tasks grappling with ambiguity and layered representations.

The \texttt{\texttt{MD-3k}} benchmark, the first of its kind with multi-layer spatial relationship labels, provides a critical platform for rigorous evaluation of multi-layer spatial understanding. \texttt{\texttt{MD-3k}} enables fine-grained analysis of depth disentanglement, pushing beyond single-layer metrics and driving progress in foundational spatial intelligence for computer vision, robotics, and AI.

Ultimately, LVP demonstrates spectral prompting as a powerful mechanism to unlock zero-shot multi-layer spatial understanding from existing models. This paradigm shift towards spectral prompting and multi-hypothesis spatial foundation models opens transformative avenues for interpretable, adaptable, and robust AI. By conquering depth ambiguity, this research delivers essential tools and insights for building foundational models capable of truly understanding real-world 3D scenes.

\section{Availability and Maintenance}
\label{sec:availablility-maintenance}

To accelerate progress in multi-layer spatial understanding and spectral prompting, all code and datasets from this study are publicly released at [Anonymous Repo].  This repository provides:

\begin{itemize}
    \item \textbf{Laplacian Visual Prompting (LVP) code}.  Ready-to-use implementation for spectral prompting.
    \item \textbf{\texttt{\texttt{MD-3k}} benchmark}. The first multi-layer spatial relationship dataset for rigorous evaluation.
    \item \textbf{Evaluation suite and baselines}. Code to reproduce and extend our experimental results.
    \item \textbf{Reproduction guide}.  Step-by-step instructions for full experiment replication.
\end{itemize}

We are committed to the sustained accessibility and usability of these resources, empowering the community to build upon this foundation and drive future innovations in ambiguity-free spatial AI. We actively encourage researchers to leverage \texttt{\texttt{MD-3k}} and LVP to advance the state-of-the-art in spatial understanding.

\section{License}\label{sec:license}
The \texttt{\texttt{MD-3k}} benchmark and the Laplacian Visual Prompting code are released under the Apache License 2.0. %Please refer to [Anonymous Repo URL will be added upon acceptance] for the full license details.

\section{Public Resources Used}
\label{sec:public-assets}

We acknowledge the following public resources used in this work:

\begin{itemize}

\item Depth-Anything-v2\footnote{\url{https://github.com/DepthAnything/Depth-Anything-V2}} \dotfill Apache-2.0+CC-BY-NC-4.0
\item Depth-Anything\footnote{\url{https://github.com/LiheYoung/Depth-Anything}.} \dotfill Apache-2.0

\item DPT\footnote{\url{https://github.com/isl-org/DPT}.} \dotfill MIT

\item ZoeDepth\footnote{\url{https://github.com/isl-org/ZoeDepth}.} \dotfill MIT

\item Marigold\footnote{\url{https://github.com/prs-eth/Marigold}.} \dotfill Apache-2.0

\item GeoWizard\footnote{\url{https://github.com/fuxiao0719/GeoWizard}.} \dotfill CC BY 4.0

\item Diffusers\footnote{\url{https://github.com/huggingface/diffusers/tree/main}} \dotfill Apache-2.0

\item Video Depth Anything\footnote{\url{https://github.com/DepthAnything/Video-Depth-Anything}} \dotfill Apache-2.0

\end{itemize}

\begin{figure*}[t!]
\centering \setlength{\abovecaptionskip}{0.1cm}
\includegraphics[width=\textwidth]{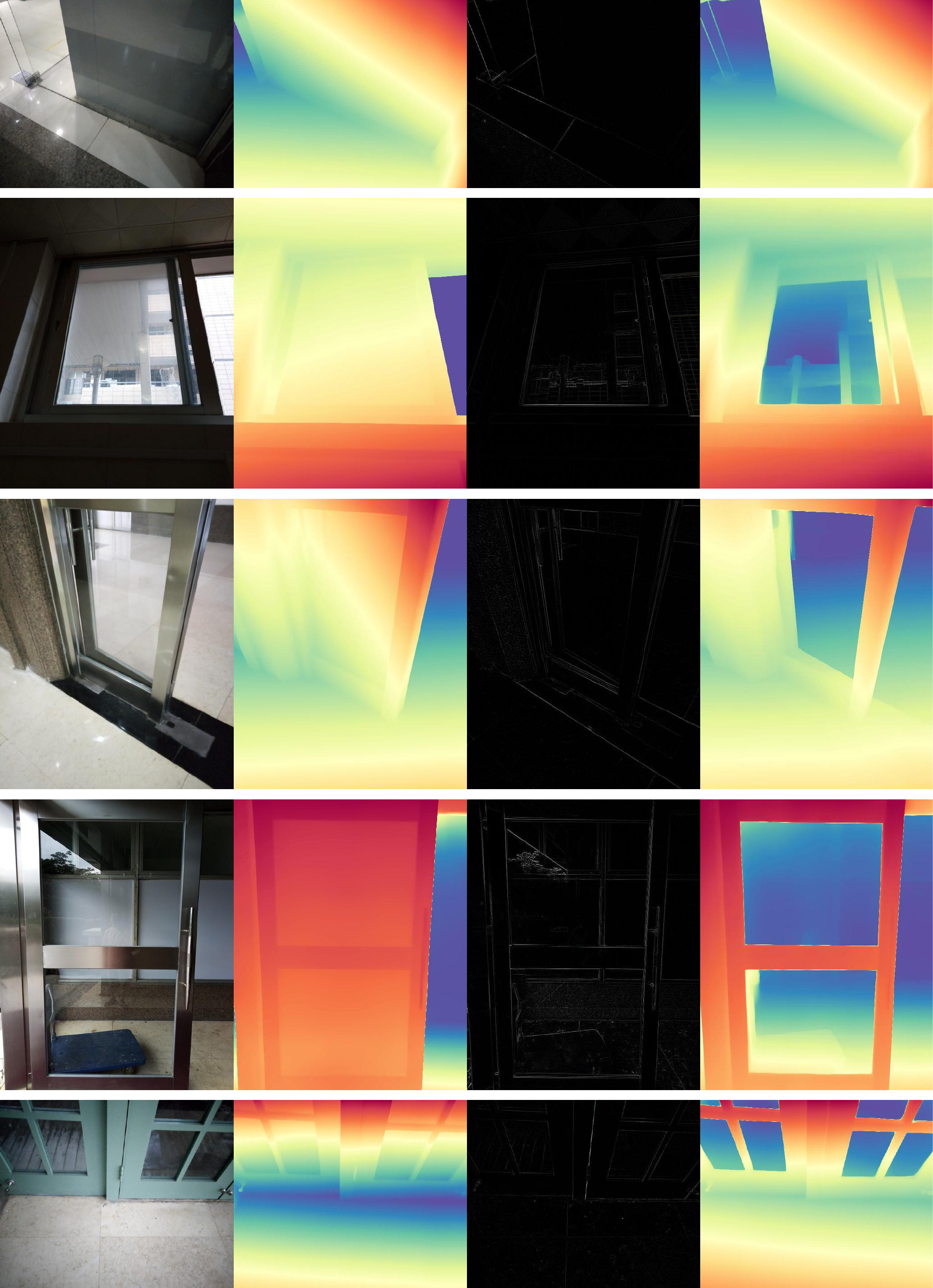}
\caption{\textbf{LVP-empowered multi-layer depth.} Each case includes an RGB image with its depth, and the Laplacian with its depth.}
\label{fig:hidden_depth_2_1}
\end{figure*}

\begin{figure*}[t!]
\centering \setlength{\abovecaptionskip}{0.1cm}
\includegraphics[width=\textwidth]{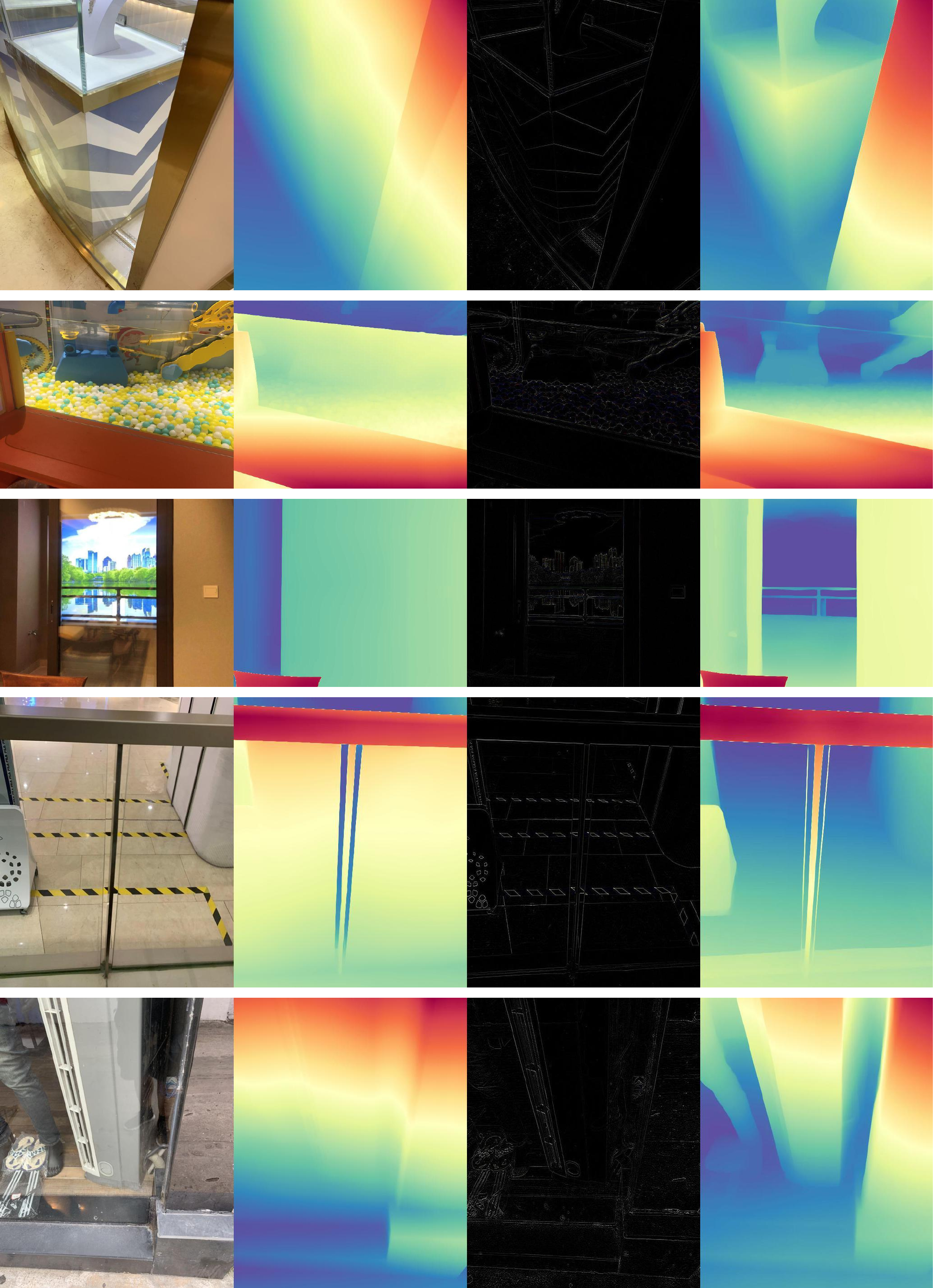}
\caption{\textbf{LVP-empowered multi-layer depth.} Each case includes an RGB image with its depth, and the Laplacian with its depth.}
\label{fig:hidden_depth_2_2}
\end{figure*}

\begin{figure*}[t!]
\centering \setlength{\abovecaptionskip}{0.1cm}
\includegraphics[width=\textwidth]{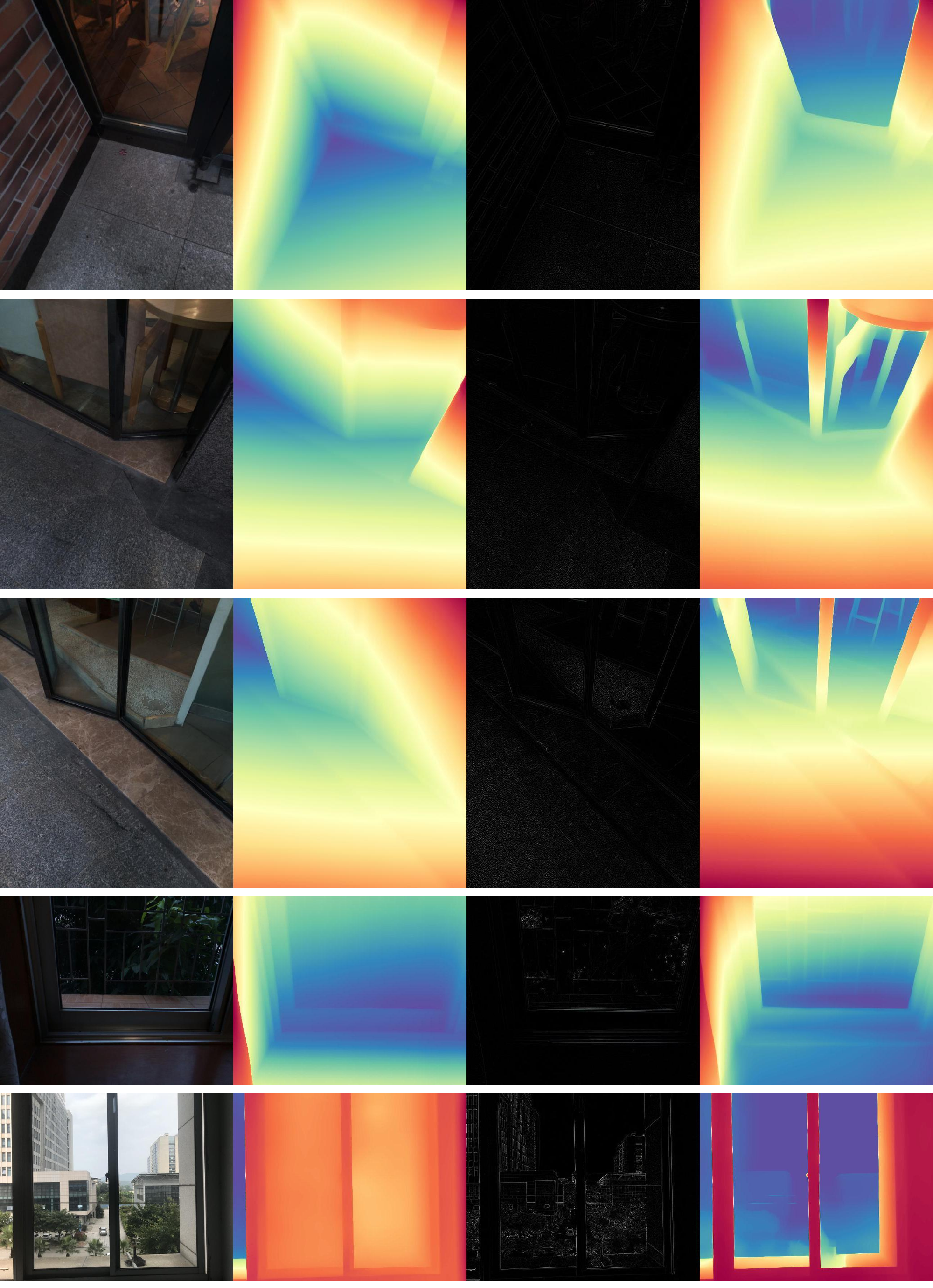}
\caption{\textbf{LVP-empowered multi-layer depth.} Each case includes an RGB image with its depth, and the Laplacian with its depth.}
\label{fig:hidden_depth_2_3}
\end{figure*}

\begin{figure*}[t!]
\centering \setlength{\abovecaptionskip}{0.1cm}
\includegraphics[width=\textwidth]{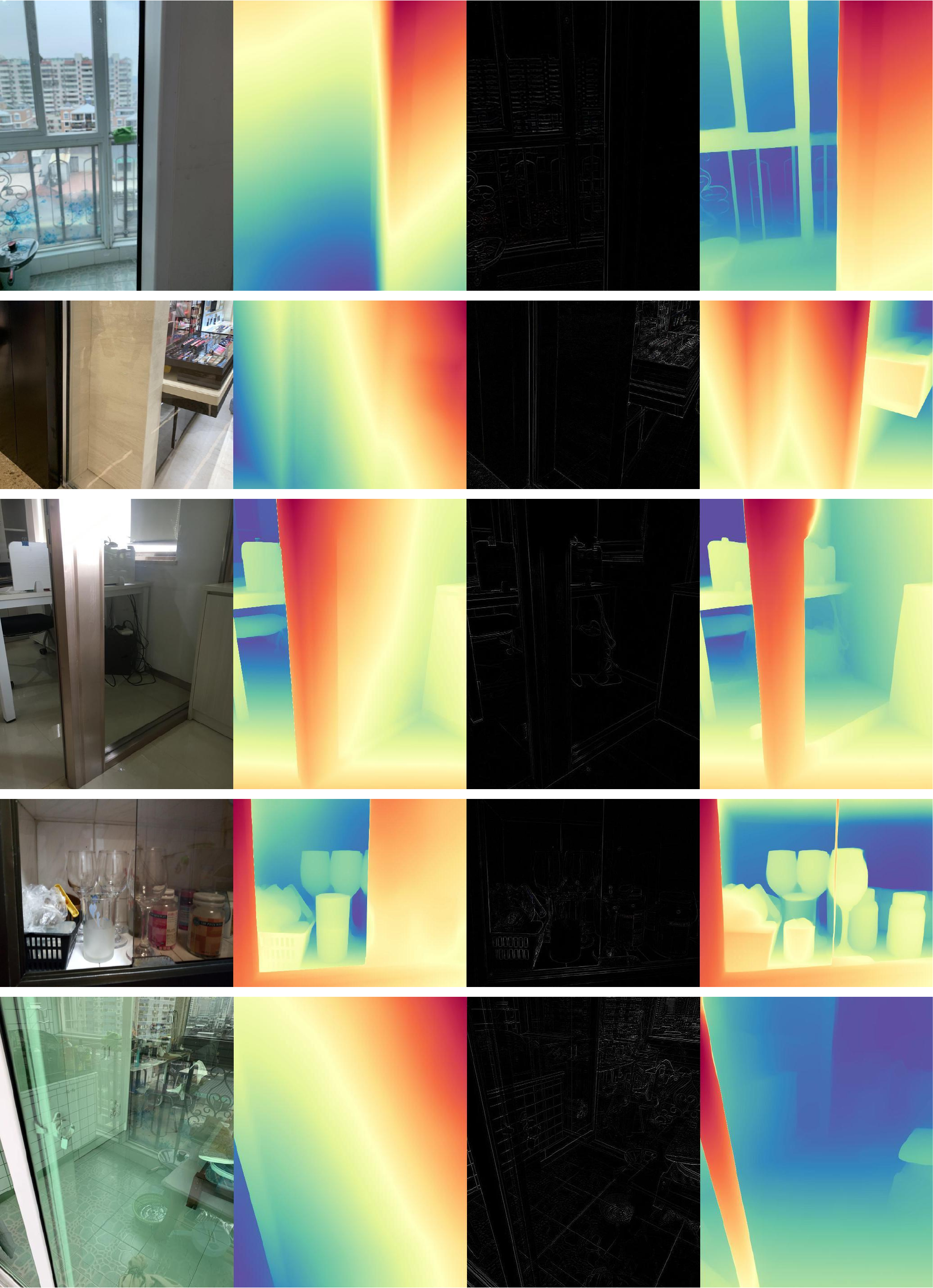}
\caption{\textbf{LVP-empowered multi-layer depth.} Each case includes an RGB image with its depth, and the Laplacian with its depth.}
\label{fig:hidden_depth_2_4}
\end{figure*}

\begin{figure*}[t!]
\centering \setlength{\abovecaptionskip}{0.1cm}
\includegraphics[width=\textwidth]{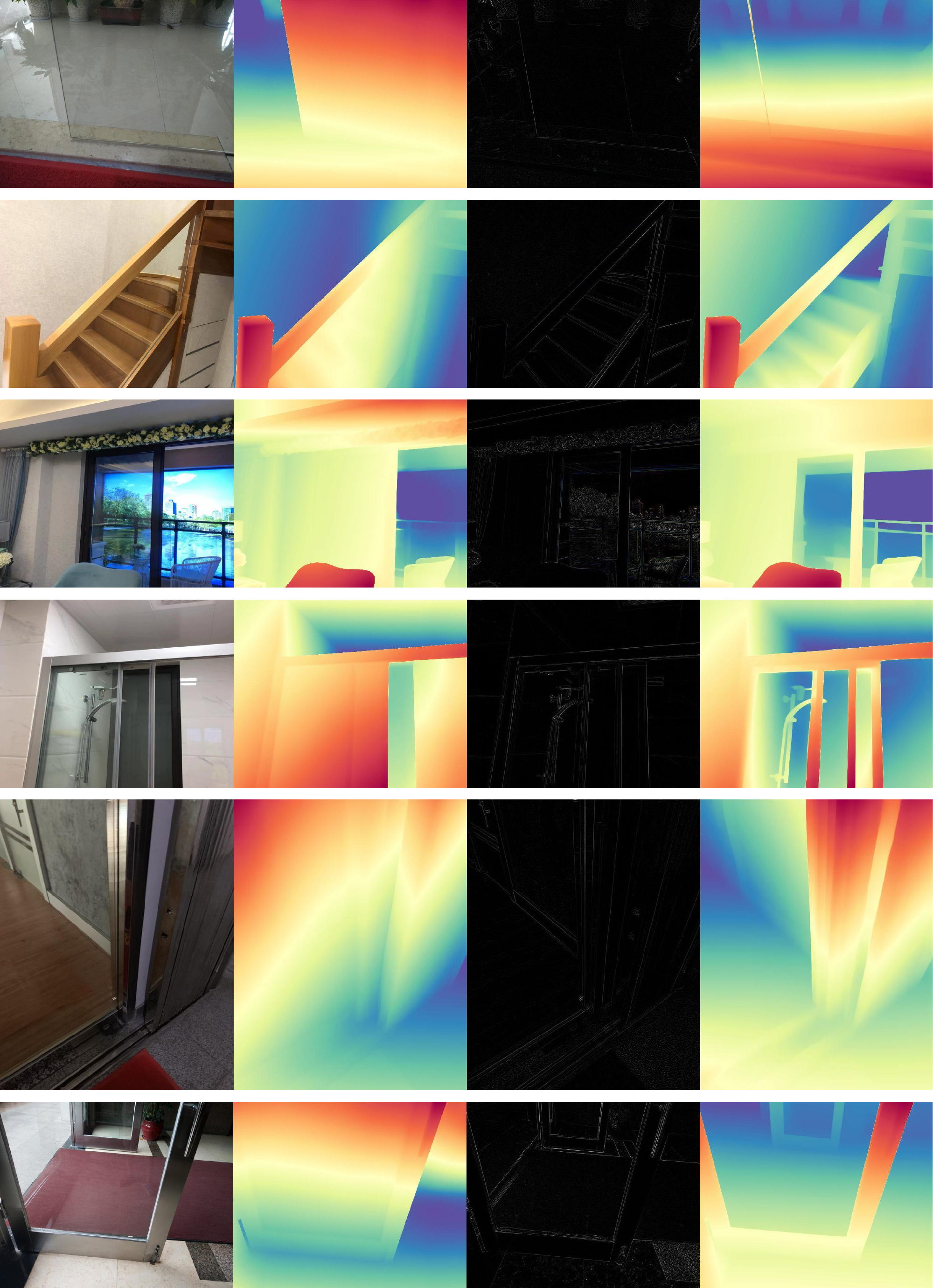}
\caption{\textbf{LVP-empowered multi-layer depth.} Each case includes an RGB image with its depth, and the Laplacian with its depth.}
\label{fig:hidden_depth_2_5}
\end{figure*}

\begin{figure*}[t!]
\centering \setlength{\abovecaptionskip}{0.1cm}
\includegraphics[width=\textwidth]{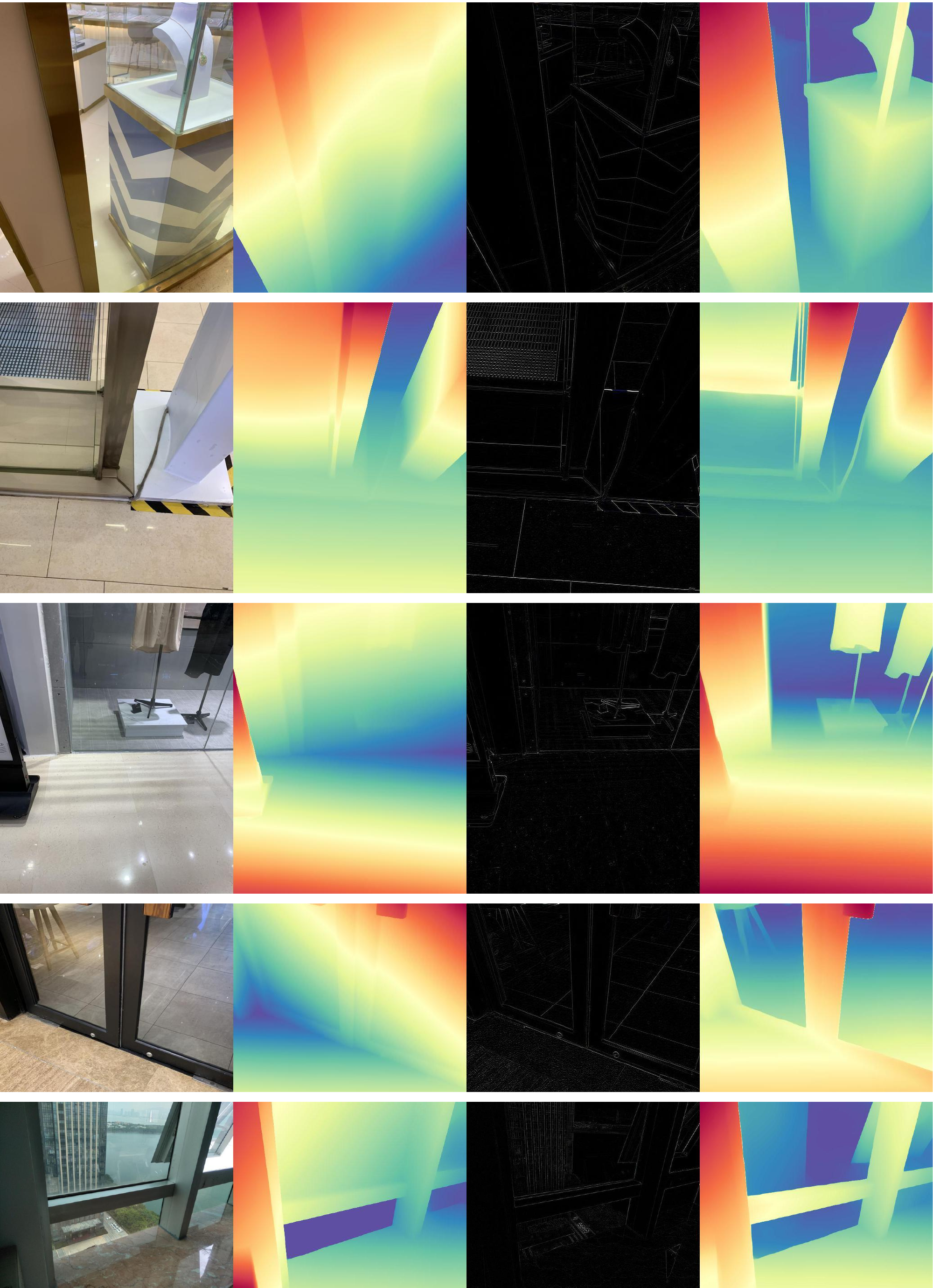}
\caption{\textbf{LVP-empowered multi-layer depth.} Each case includes an RGB image with its depth, and the Laplacian with its depth.}
\label{fig:hidden_depth_2_6}\vspace{-2mm}
\end{figure*}

\begin{figure*}[t!] %\vspace{-5mm}
    \centering\setlength{\abovecaptionskip}{0.1cm}
    \includegraphics[width=\textwidth]{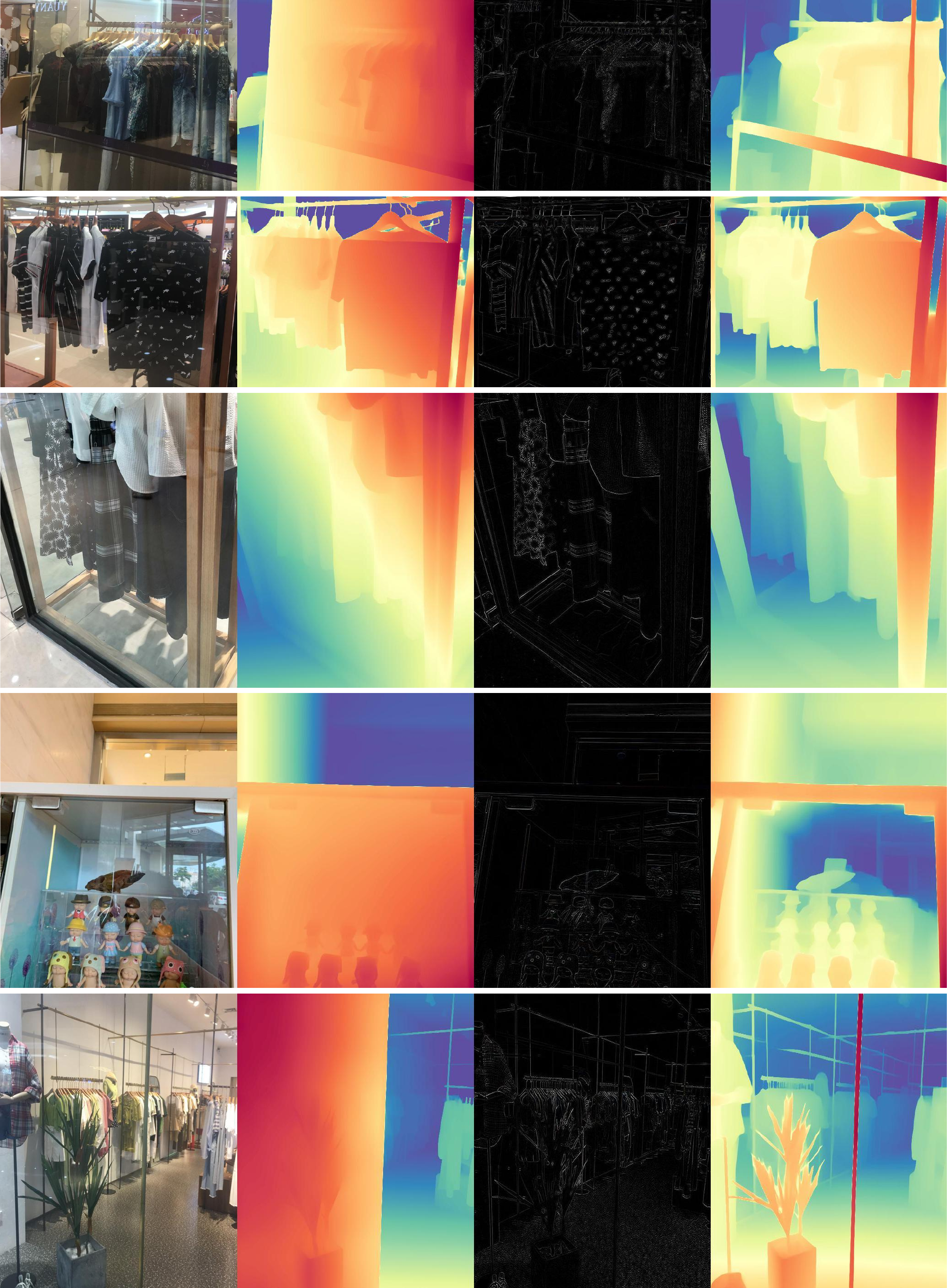}
\caption{\textbf{\textcolor[HTML]{DB4437}{Failure cases} of multi-layer depth estimation via Laplacian Visual Prompting.} Each case shows the RGB input, the estimated depth via RGB, the Laplacian input, and the estimated \textit{depth} via Laplacian input. }
    \label{fig:failure_cases}\vspace{-2mm}
\end{figure*}

\begin{figure*}[t!]
\centering  \setlength{\abovecaptionskip}{0.1cm}
\includegraphics[width=\textwidth]{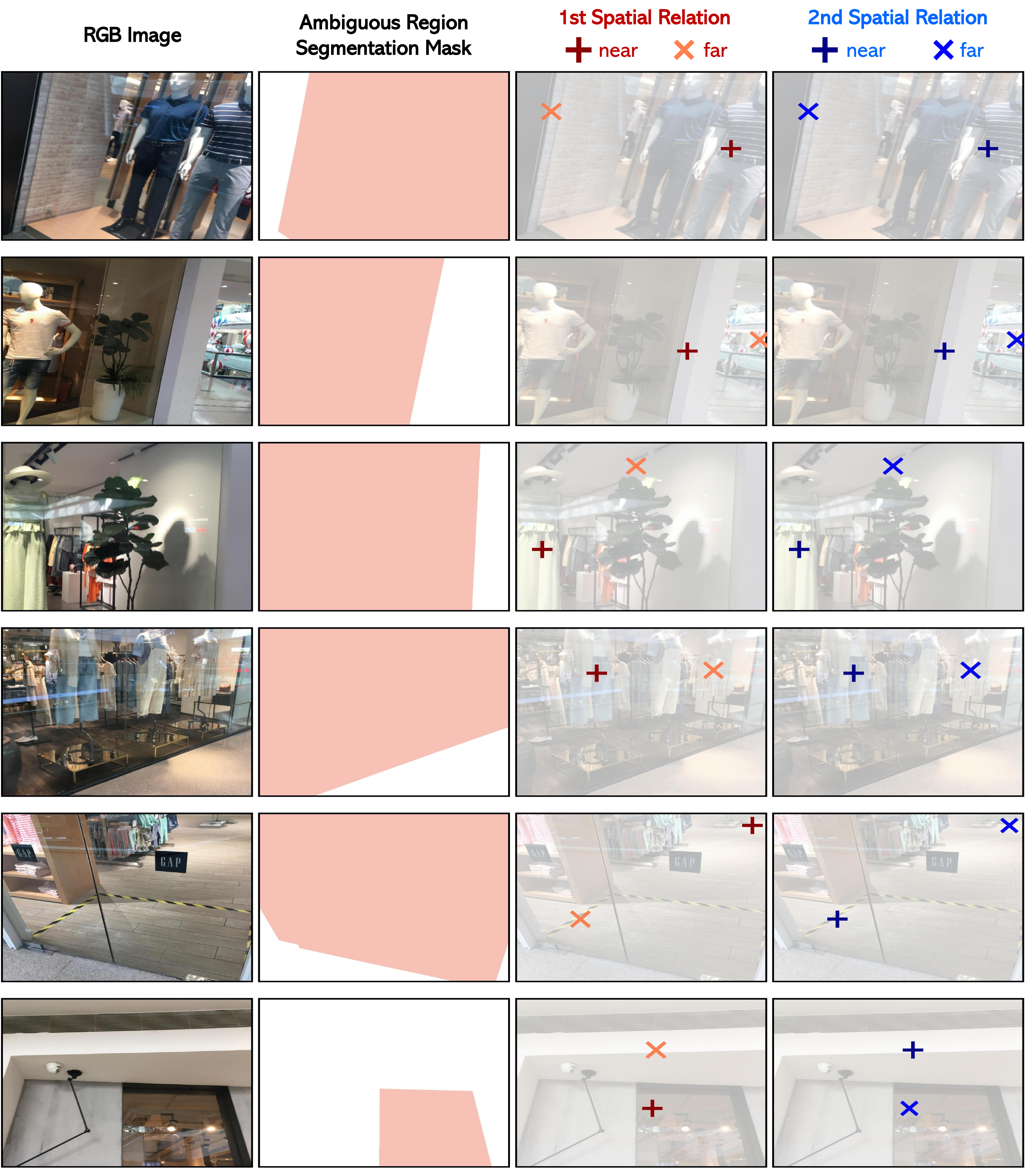}
 \caption{\textbf{MD-3k benchmark for evaluating multi-layer spatial relationships.}  Example images with annotated sparse point pairs are shown, illustrating ambiguous regions and relative depth relationships.  The first and second spatial relation columns show ground truth annotations for near/far relationships between layers, using red and blue markers, respectively.}
\label{fig:md3k_case1}\vspace{-2mm}
\end{figure*}

\begin{figure*}[t!]
\centering \setlength{\abovecaptionskip}{0.2cm}
\includegraphics[width=\textwidth]{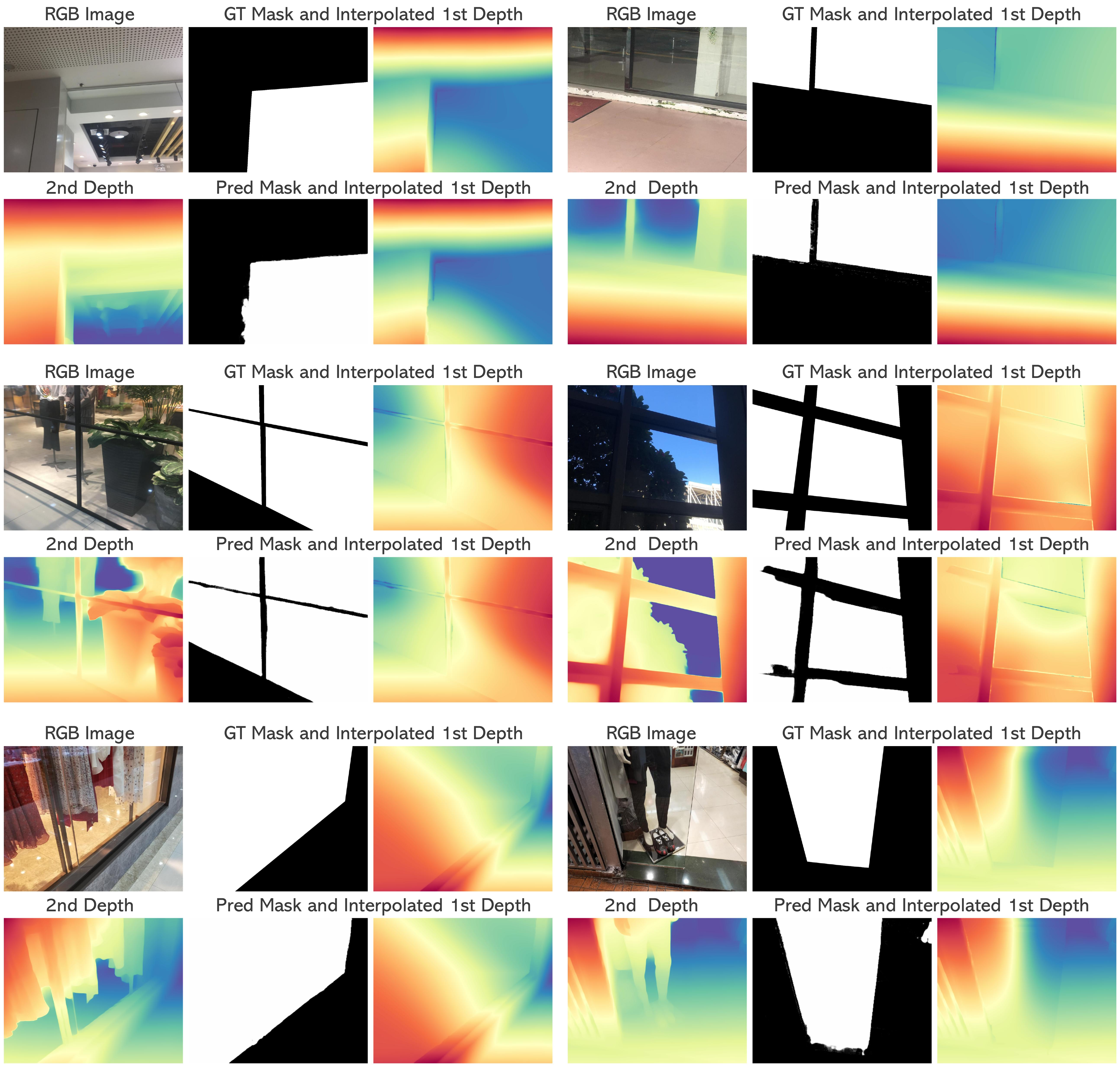}
\caption{\textbf{Multi-layer depth with extra semantic prior ({successful} {cases}).}}
\label{fig:depth_seg_good}%\vspace{-2mm}
\end{figure*}

\begin{figure*}[t!]
\centering \setlength{\abovecaptionskip}{0.2cm}
\includegraphics[width=\textwidth]{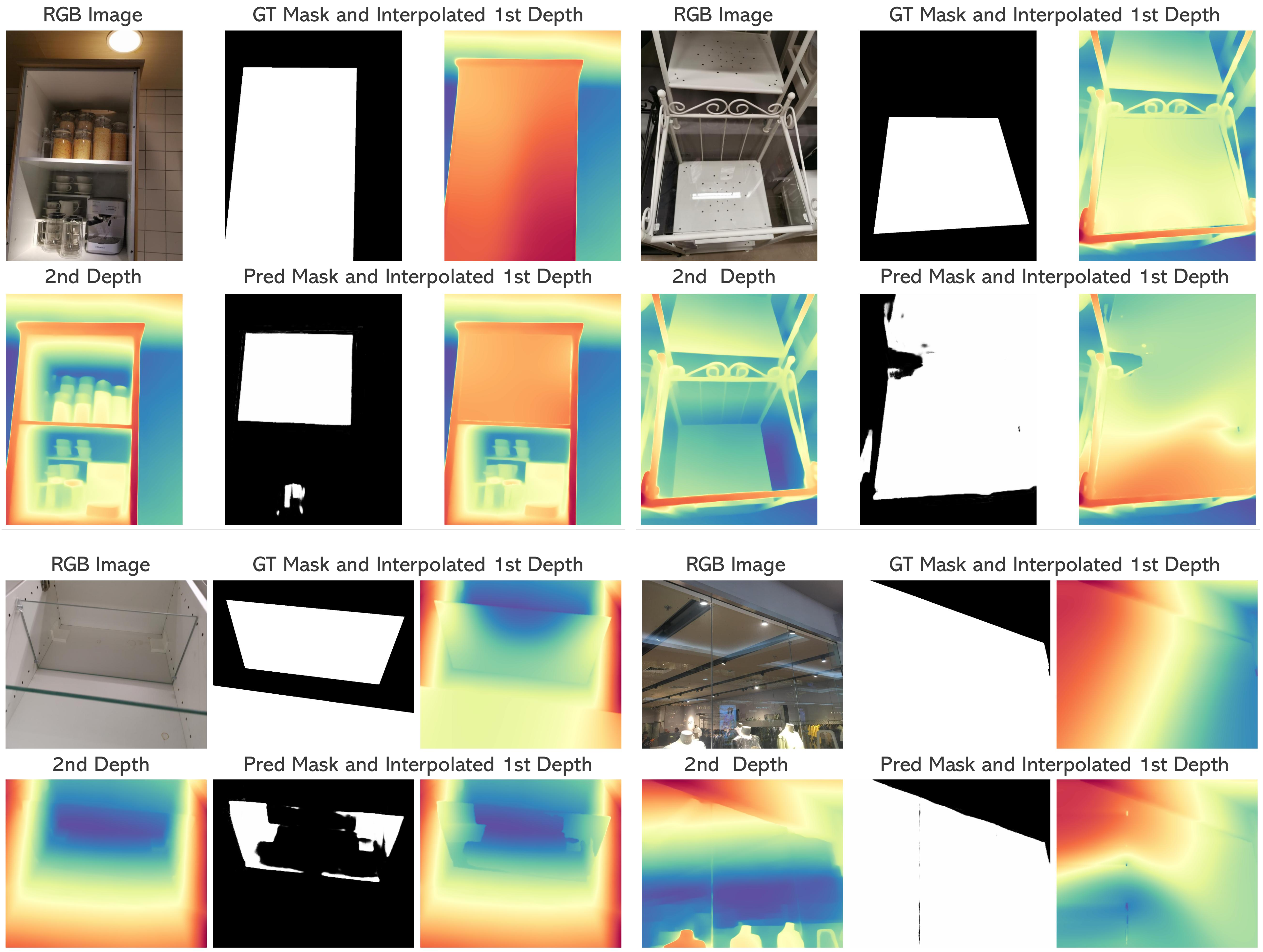}
\caption{\textbf{Multi-layer depth with extra semantic prior ({failure} {cases}).}}
\label{fig:depth_seg_bad}%\vspace{-2mm}
\end{figure*}

\clearpage
{
    \small
    \bibliographystyle{ieeenat_fullname}
    \bibliography{main}
}

%\clearpage
%\input{sec/X_suppl_new}


\end{document}